%% file: situation17.tex
\documentclass[10pt,twocolumn,letterpaper]{article}

\usepackage{iccv}
\usepackage{times}
\usepackage{epsfig}
\usepackage{graphicx}
\graphicspath{{figures/}}
\usepackage{amsmath}
\usepackage{amssymb}

\usepackage{balance}
\usepackage{booktabs}
\usepackage{multirow}
\usepackage{epsfig}
\usepackage{epstopdf}
\usepackage{color}
\usepackage{lipsum}

\newcommand*\rot{\rotatebox{90}}

\usepackage[pagebackref=true,breaklinks=true,letterpaper=true,colorlinks,bookmarks=false]{hyperref}

\iccvfinalcopy 

\def\iccvPaperID{1857} 

\ificcvfinal\fi
\begin{document}

\title{Situation Recognition with Graph Neural Networks}

\author{
Ruiyu Li$^1$, \hspace{2mm}Makarand Tapaswi$^2$, \hspace{2mm}Renjie Liao$^{2,4}$, \hspace{2mm}Jiaya Jia$^{1,3}$, \hspace{2mm}Raquel Urtasun$^{2,4,5}$, \hspace{2mm}Sanja Fidler$^{2,5}$
\vspace{0.3cm}\\
$^1$The Chinese University of Hong Kong, $^2$University of Toronto, $^3$Youtu Lab, Tencent\\$^4$Uber Advanced Technologies Group, $^5$Vector Institute
\\
\texttt{\footnotesize ryli@cse.cuhk.edu.hk,
\{makarand,rjliao,urtasun,fidler\}@cs.toronto.edu, leojia9@gmail.com }}

\maketitle

\begin{abstract}
We address the problem of recognizing situations in images.
Given an image, the task is to predict the most salient verb (action), and fill its semantic roles such as who is performing the action, what is the source and target of the action, \etc.
Different verbs have different roles (\eg~\texttt{attacking} has \texttt{weapon}), and each role can take on many possible values (nouns).
We propose a model based on Graph Neural Networks that allows us to efficiently capture joint dependencies between roles using neural networks defined on a graph.
Experiments with different graph connectivities show that our approach that propagates information between roles significantly outperforms existing work, as well as multiple baselines.
We obtain roughly 3-5\% improvement over previous work in predicting the full situation.
We also provide a thorough qualitative analysis of our model and influence of different roles in the verbs.
\end{abstract}

\input{main_sections/introduction}
\input{main_sections/relwork}
\input{main_sections/models}
\input{main_sections/evaluation}

\input{main_sections/conclusion}

\begin{small}
\section*{Acknowledgements}
\vspace{-2mm}

This work is in part supported by a grant from the Research Grants Council of the Hong
Kong SAR (project No. 413113). We also acknowledge support from NSERC, and GPU donations from NVIDIA.
\end{small}

{\small
\bibliographystyle{ieee}
\bibliography{situation17}
}

\newpage
\appendix
\input{main_sections/appendix}

\end{document}

%% file: main_sections/introduction.tex

\vspace{-3mm}
\section{Introduction}
\label{sec:intro}

Object~\cite{resnet,imagenet,simonyan2015verydeep}, action~\cite{simonyan2014two,wang2011action}, and scene classification~\cite{zhou2016places365,zhou2014places} have come a long way, with performance in some of these tasks almost reaching human agreement.
However, in many real world applications such as robotics we need a much more detailed understanding of the scene.
For example, knowing that an image depicts a \texttt{repairing} action is not sufficient to understand what is really happening in the scene.
We thus need additional information such as the person repairing the house, and the tool that is used.

\begin{figure}[t]
\centering
\includegraphics[width=0.74\linewidth]{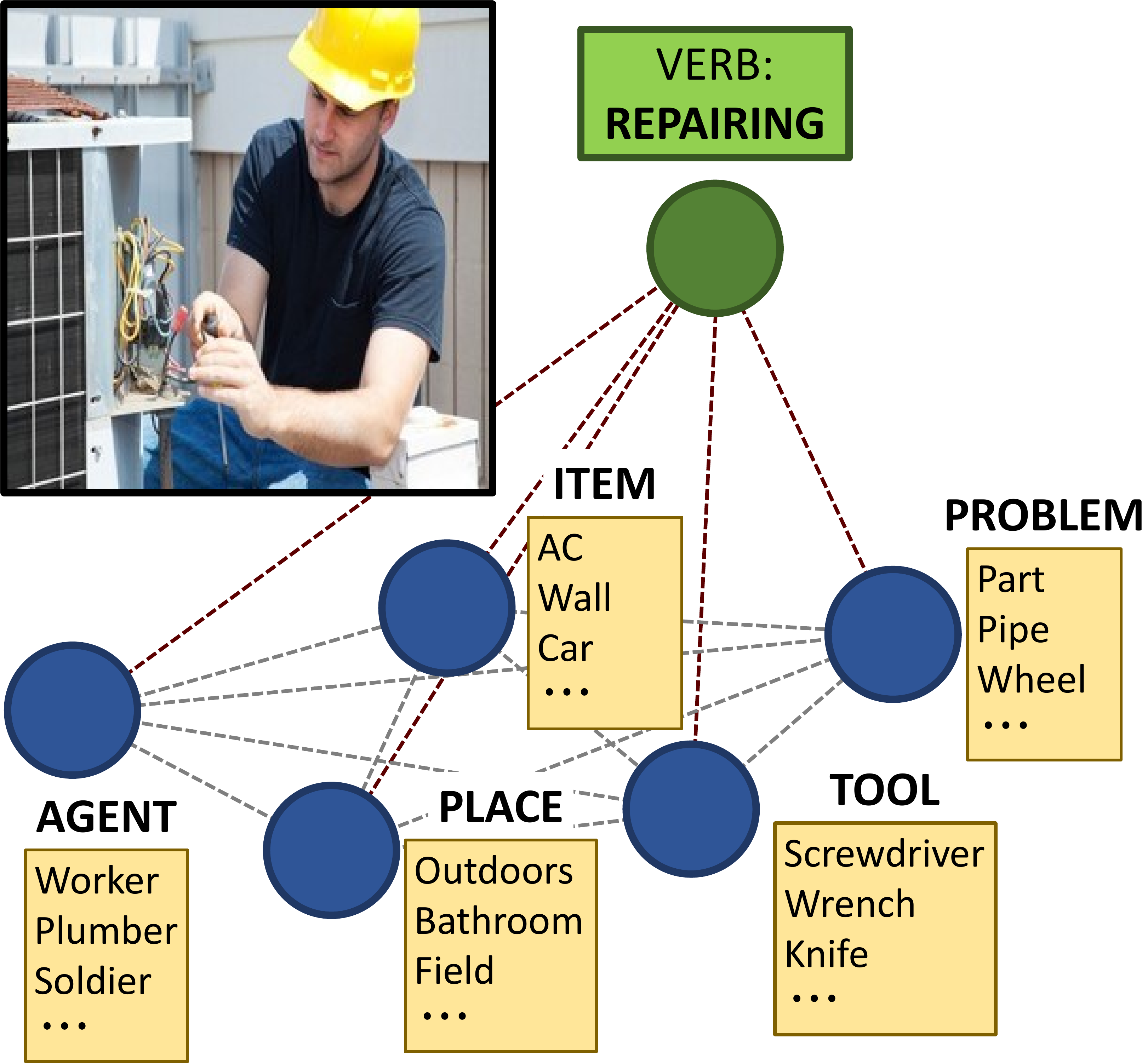}
\caption{Understanding an image involves more than just predicting the most salient action.
We need to know who is performing this action, what tools (s)he may be using,~\etc.
Situation recognition is a structured prediction task that aims to predict the verb and its \emph{frame} that consists of multiple role-noun pairs.
The figure shows a glimpse of our model that uses a graph to model dependencies between the verb and its roles.
}
\vspace{-0.2cm}
\label{fig:intro}
\end{figure}

\begin{figure}[t]
\centering
\includegraphics[width=0.83\linewidth]{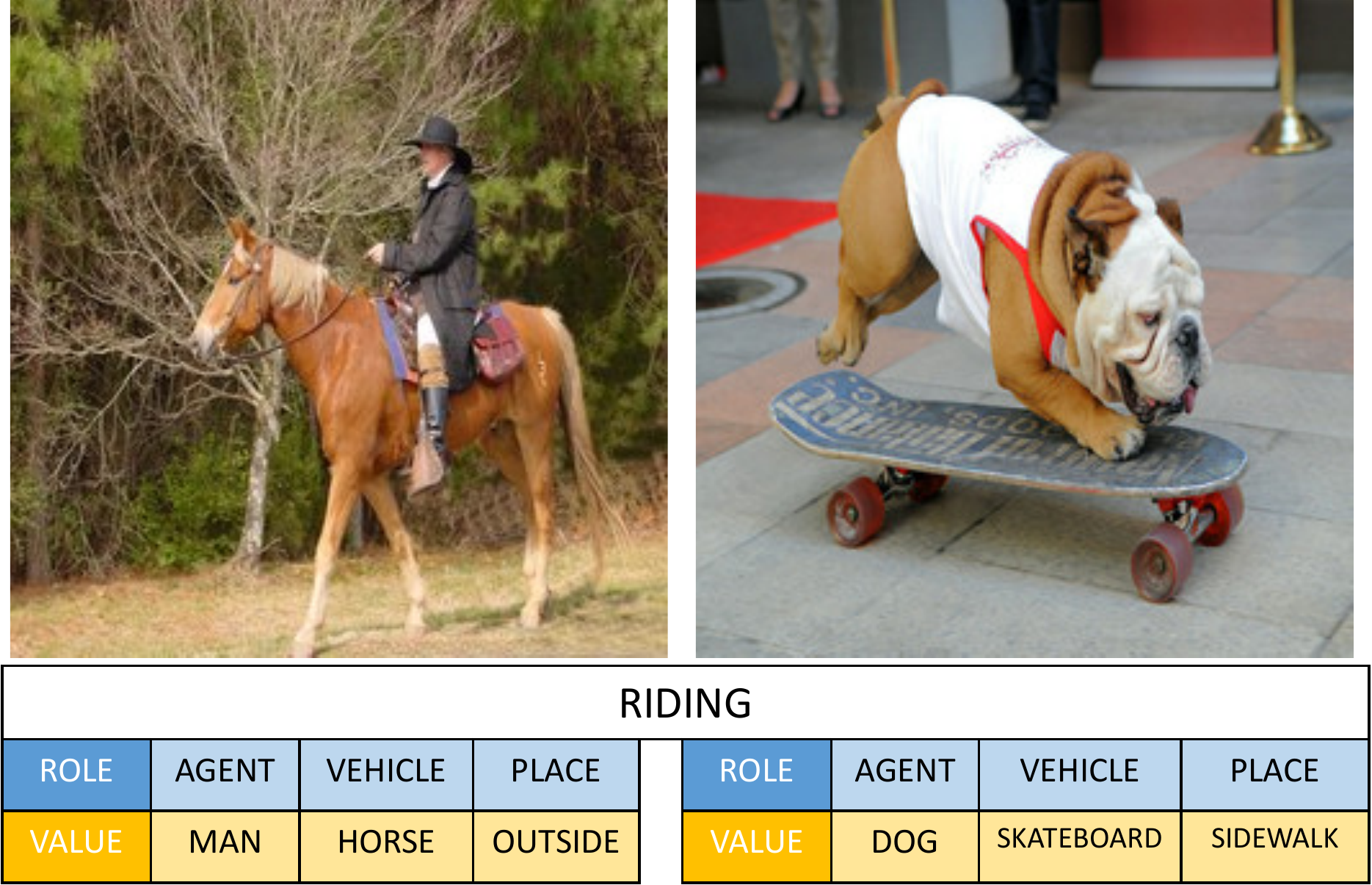}
\caption{Images corresponding to the same verb can be quite different in their content involving verb roles.
This makes situation recognition difficult.}
\label{fig:examples}
\vspace{-0.3cm}
\end{figure}

Several datasets have recently been collected for such detailed understanding~\cite{krishna2016genome,lukrishna2016vrd,yatskar2016imsitu}.
In~\cite{krishna2016genome}, the Visual Genome dataset was built containing detailed relationships between objects.
A subset of the scenes were further annotated with \emph{scene graphs}~\cite{johnson2015retrieval} to capture both unary (\eg~attributes) and pairwise (\eg~relative spatial info) object relationships.
Recently, Yatskar et al.~\cite{yatskar2016imsitu} extended this idea to actions by labeling action \emph{frames} where a frame consists of a fixed set of roles that define the action.
Fig.~\ref{fig:intro} shows a frame for action \texttt{repairing}.
The challenge then consists of assigning values (nouns) to these roles based on the image content.
The number of different role types, their possible values, as well as the number of actions are very large, making it a very challenging prediction task.
As shown in Fig.~\ref{fig:examples}, the same verb can appear in very different image contexts, and nouns that fill the roles are vastly different.

In~\cite{yatskar2016imsitu}, the authors proposed a Conditional Random Field (CRF) to model dependencies between verb-role-noun pairs.
In particular, a neural network was trained in an end-to-end fashion to both, predict the unary potentials for verbs and nouns, and to perform inference in the CRF.
While their model captured the dependency between the verb and role-noun pairs, dependencies between the roles were not modeled explicitly.

In this paper, we aim to jointly reason about verbs and their roles using a \emph{Graph Neural Network} (GNN), a generalization of graphical models to neural networks.
A GNN defines observation and output at each node in the graph, and propagates messages along the edges in a recurrent manner.
In particular, we exploit the GNNs to also model dependencies between roles and predict a consistent structured output.
We explore different connectivity structures among the role nodes, and show that our approach significantly improves performance over existing work.
In addition, we compare with strong baseline methods using Recurrent Neural Networks (RNNs) that have been shown to work well on joint prediction tasks, such as semantic~\cite{zheng2015_CRF-RNN} and object instance~\cite{polyrnn} segmentation, as well as on group activity recognition~\cite{deng2016structmach}.
We also visualize the learned models to further investigate dependencies between roles.

%% file: main_sections/relwork.tex
\vspace{-0mm}
\section{Related Work}
\label{sec:relwork}


Situation recognition generalizes action recognition to include actors, objects, and location in the activity.
There has been work to combine activity recognition with scene or object labels~\cite{delaitre2010,gupta2008,yao2010,yao2011}.
In~\cite{gupta2015visual,ronchi2015describing}, visual semantic role labeling tasks were proposed where datasets are built to study action along with localization of people and objects.
In another line of work, Yatskar~\etal~\cite{yatskar2016imsitu} created the \emph{imSitu} dataset that uses linguistic resources from FrameNet~\cite{framenet} and WordNet~\cite{wordnet} to associate images not only with verbs, but also with specific role-noun pairs that describe the verb with more details.
As a baseline approach, in \cite{yatskar2016imsitu}, a Conditional Random Field (CRF) jointly models prediction of the verb and verb-role-noun triplets.
Further, considering that the large output space and sparse training data could be problematic, a tensor composition function was used \cite{yatskar2016commonly} to share nouns across different roles.
The authors also proposed to augment the training data by searching images using query phrases built from the structured situation.

Different from these methods, our work focuses on explicitly modeling dependencies between roles for each verb through the use of different neural architectures.

\vspace{-3mm}
\paragraph{Understanding Images.}
There is a surge of interest in joint vision and language tasks in recent years.
Visual Question Answering in images and videos~\cite{antol2015vqa,tapaswi2016movieqa} aims to answer questions related to image or video content.
In image captioning~\cite{karpathy2015devise,vinyals2015showtell,xu2015showattendtell,Ling17}, a natural language sentence is generated to describe the image.
Approaches for these tasks often use the CNN-RNN pipelines to provide a caption, or a correct answer to a specific question. Dependencies between verbs and nouns are typically being implicitly learned with the RNN. 
An alternative is to list all important objects with their attributes and relationships.
Johnson~\etal~\cite{johnson2015retrieval} created \emph{scene graphs}, which are being used for visual relationship detection~\cite{lukrishna2016vrd,plummer2017phrase,zhang2017visual} tasks. In~\cite{LinBMVC15}, the authors exploit scene graphs to generate image captions. 

In Natural Language Processing (NLP), semantic role labeling~\cite{furstenau2009graph,jurafsky_martin,kingsbury2002,roth2016neural,yang2016grounding,zhou2015end2end} involves annotating a sentence with thematic or semantic roles.
Building upon resources from NLP, and leveraging collections such as FrameNet~\cite{framenet} and WordNet~\cite{wordnet}, visual semantic role labeling, or situation recognition, aims to interpret details for one particular action with verb-role-noun pairs.

\vspace{-3mm}
\paragraph{Graph Neural Networks.}
There are a few different ways for applying neural networks to graph-structured data.
We divide them into two categories.
The first group defines convolutions on graphs.
Approaches like~\cite{bruna2013spectral,defferrard2016convolutional,kipf2016semi} utilized the graph Laplacian and applied CNNs to spectral domain.
Differently, Duvenaud et al.~\cite{duvenaud2015convolutional} designed a special hash function such that a CNN can be used on the original graphs.

The second group applies feed-forward neural networks to every node of the graph recurrently.
Information is propagated through the network by dynamically updating the hidden state of each node based on their history and incoming messages from their neighborhood.
The Graph Neural Network (GNN) proposed by~\cite{scarselli2009graph} utilized multi-layer perceptrons (MLP) to update the hidden state.
However, their learning algorithm is restrictive due to the contraction map assumption.
In the following work, the Gated Graph Neural Network (GGNN)~\cite{li2016gated} used a recurrent gating function~\cite{cho2014gru} to perform the update, and effectively learned model parameters using back-propagation through time (BPTT).

Other work~\cite{liang2016semantic,tai2015improved} designed special update functions based on the LSTM~\cite{hochreiter1997long} cell and applied the model to tree-structured or general graph data.
In \cite{marino2016more}, knowledge graphs and GGNNs are used for image classification. Here we use GGNNs for situation recognition.

%% file: main_sections/models.tex

\section{Graph-based Neural Models for Situation Recognition}
\label{sec:models}

\paragraph{Task Definition.}
Situation recognition as per the \emph{imSitu} dataset~\cite{yatskar2016imsitu} assumes a discrete set of verbs $\mathcal{V}$, nouns $\mathcal{N}$, roles $\mathcal{R}$, and frames $\mathcal{F}$.
The verb and its corresponding frame that contains roles are obtained from FrameNet~\cite{framenet}, while nouns come from WordNet~\cite{wordnet}.
Each verb $v \in \mathcal{V}$ is associated with a frame $f \in \mathcal{F}$ that contains a set of semantic roles $E_f$.
Each role $e \in E_f$ is paired with a noun value $n_e \in N \cup \{\varnothing\}$.
Here, $\varnothing$ indicates that the noun is unknown or not applicable.
A set of semantic roles and their nouns is called a realized frame, denoted as {\small$R_f = \{(e, n_e) : e \in E_f\}$}, where each role is with a noun.

Given an image, the task is to predict the structured situation $S = (v, R_f)$, specified by a verb $v \in \mathcal{V}$ and its corresponding realized frame $R_f$.
For example, as shown on the right of Fig.~\ref{fig:examples}, the verb \texttt{riding} is associated with three role-noun pairs, i.e., \texttt{\small\{agent:dog, vehicle:surfboard, place:sidewalk\}}.

\begin{figure}[t]
\vspace{-1.5mm}
\centering
\includegraphics[width=0.88\linewidth]{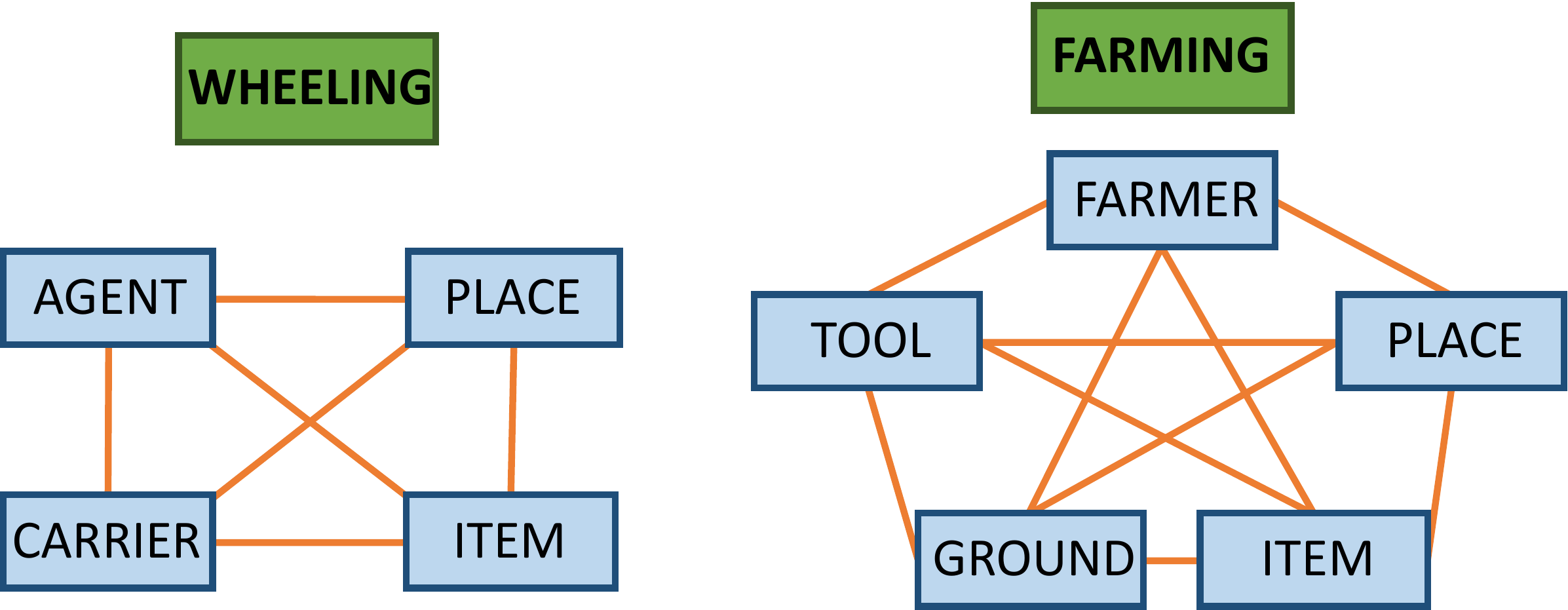}
\caption{The architecture of fully-connected roles GGNN.
The undirected edges between all roles of a verb-frame allows to fully capture the dependencies between them.}
\label{fig:fc_ggnn}
\vspace{-2mm}
\end{figure}

\subsection{Graph Neural Network}
\label{subsec:models:ggnn}

The verb and semantic roles of a situation depend on each other.
For example, in the verb \texttt{carrying}, the roles \texttt{agent} and \texttt{agent-part} are tightly linked with the \texttt{item} being carried.
Small items can be carried by hand, while heavy items may be carried on the back.
We propose modeling these dependencies through a graph $G = (\mathcal{A}, \mathcal{B})$.
The nodes in our graph $a \in \mathcal{A}$ are of two types of \emph{verb} or \emph{role}, and take unique values of $\mathcal{V}$ or $\mathcal{N}$, respectively.
Since each image in the dataset is associated with one unique verb, every graph has a single verb node.
Edges in the graph $b = (a', a)$ encode dependencies between role-role or verb-role pairs, and can be directed or undirected.
Fig.~\ref{fig:intro} shows an example of such a graph where verb and role nodes are connected to each other.

\vspace{-2mm}
\paragraph{Background.}
Modeling structure and learning representation on graphs have prior work.
Gated Graph Neural Networks (GGNNs)~\cite{li2016gated} is one approach that learns the representation of a graph, which is then used to predict node- or graph-level output.
Each node of a GGNN is associated with a hidden state vector that is updated in a recurrent fashion.
At each time step, the hidden state of a node is updated based on its history and incoming messages from its neighbors.
These updates are applied simultaneously to all nodes in the graph at each propagation step.
The hidden states after $T$ propagation steps are used to predict the output.
In contrast, a standard unrolled RNN only moves information in one direction and updates one ``node'' per time step.

\vspace{-2mm}
\paragraph{GGNN for Situation Recognition.}
We adopt the GGNN framework to recognize situations in images.
Each image $i$ is associated with one verb $v$ that corresponds to a frame $f$ with a set of roles $E_f$.
We instantiate a graph $\mathcal{G}_f$ for each image that consists of one verb node, and $|E_f|$ (number of roles associated with the frame) role nodes.
To capture the dependency between roles to the full extent, we propose creating undirected edges between all pairs of roles.
Fig.~\ref{fig:fc_ggnn} shows two example graph structures of this type. We explore other edge configurations in the evaluation.

To initialize the hidden states for each node, we use features derived from the image. 
In particular, for every image $i$, we compute representations $\phi_v(i)$ and $\phi_n(i)$ using the penultimate fully-connected layer of two convolutional neural network (CNN) pre-trained to predict verbs and nouns, respectively.
We initialize the hidden states $h \in \mathbb{R}^D$ of the verb node $a_v$ and role node $a_e$ as
\vspace{-0.1cm}
\begin{align}
\label{eq:init-verb}
h_{a_v}^0 &= g(W_{iv} \phi_v(i)) \\
\label{eq:init-noun}
h_{a_e}^0 &= g(W_{in} \phi_n(i) \odot W_e e \odot W_v \hat{v}) \, ,
\vspace{-0.2cm}
\end{align}
where $\hat{v} \in \{0, 1\}^{|\mathcal{V}|}$ corresponds to a one-hot encoding of the predicted verb and $e \in \{0, 1\}^{|\mathcal{R}|}$ is a one-hot encoding of the role that the node $a_e$ corresponds to.
$W_v \in \mathbb{R}^{D \times |\mathcal{V}|}$ is the verb embedding matrix, and $W_e \in \mathbb{R}^{D \times |\mathcal{R}|}$ is the role embedding matrix.
$W_{iv}$ and $W_{in}$ are parameters that transform image features to the space of hidden representations.
$\odot$ corresponds to element-wise multiplication, and $g(\cdot)$ is a non-linear function such as $\tanh(\cdot)$ or ReLU ($g(x) = \max(0, x)$).
We normalize the initialized hidden states to unit-norm prior to propagation.

For any node $a$, at each time step, the aggregation of incoming messages at time $t$ is determined by the hidden states of its neighbors $a'$:
\vspace{-0.1cm}
\begin{equation}
\label{eq:hidden}
x_a^t = \sum_{(a', a) \in \mathcal{B}} W_p h_{a'}^{t-1} + b_p \, .
\vspace{-0.2cm}
\end{equation}
Note that we use a shared linear layer of weights $W_p$ and biases $b_p$ to compute incoming messages across all nodes.

After aggregating the messages, the hidden state of the node is updated through a gating mechanism similar to the Gated Recurrent Unit~\cite{cho2014gru,li2016gated} as follows:
\vspace{-0.1cm}
\begin{align}
\label{eq:propagation}
z_a^t &= \sigma(W_z x_a^t + U_z h_a^{t-1} + b_z) \, , \nonumber \\
r_a^t &= \sigma(W_r x_a^t + U_r h_a^{t-1} + b_r) \, , \nonumber \\
\tilde{h}_a^t &= \tanh ( W_h x_a^{(t)} + U_h (r_a^t \odot h_a^{t-1}) + b_h) \, , \nonumber\\
h_a^{t} &= (1 - z_a^t) \odot h_a^{t-1} + z_a^t \odot \tilde{h}_a^t \quad.
\end{align}
This allows each node to softly combine the influence of the aggregated incoming message and its own memory.
$W_z$, $U_z$, $b_z$, $W_r$, $U_r$, $b_r$, $W_h$, $U_h$, and $b_h$ are the weights and biases of the update function.

\vspace{-2mm}
\paragraph{Output and Learning.}
We run $T$ propagation steps.
After propagation, we extract node-level outputs from GGNN to predict the verb and nouns.
Specifically, for each image, we predict the verb and a set of nouns for each role associated with the verb frame using a softmax layer:
\begin{align}
\label{eq:pred-verb}
p_v &= \sigma(W_{hv} h_{a_v} + b_{hv}) \\
\label{eq:pred-noun}
p_{e:n} &= \sigma(W_{hn} h_{a_e} + b_{hn}) \, .
\end{align}
Note that the softmax function $\sigma$ is applied across the class space for verbs $\mathcal{V}$ and nouns $\mathcal{N}$.
$p_{e:n}$ can be treated as the probability of assigning noun $n$ to role $e$.

\begin{figure}[t]
\centering
\includegraphics[width=0.8\linewidth]{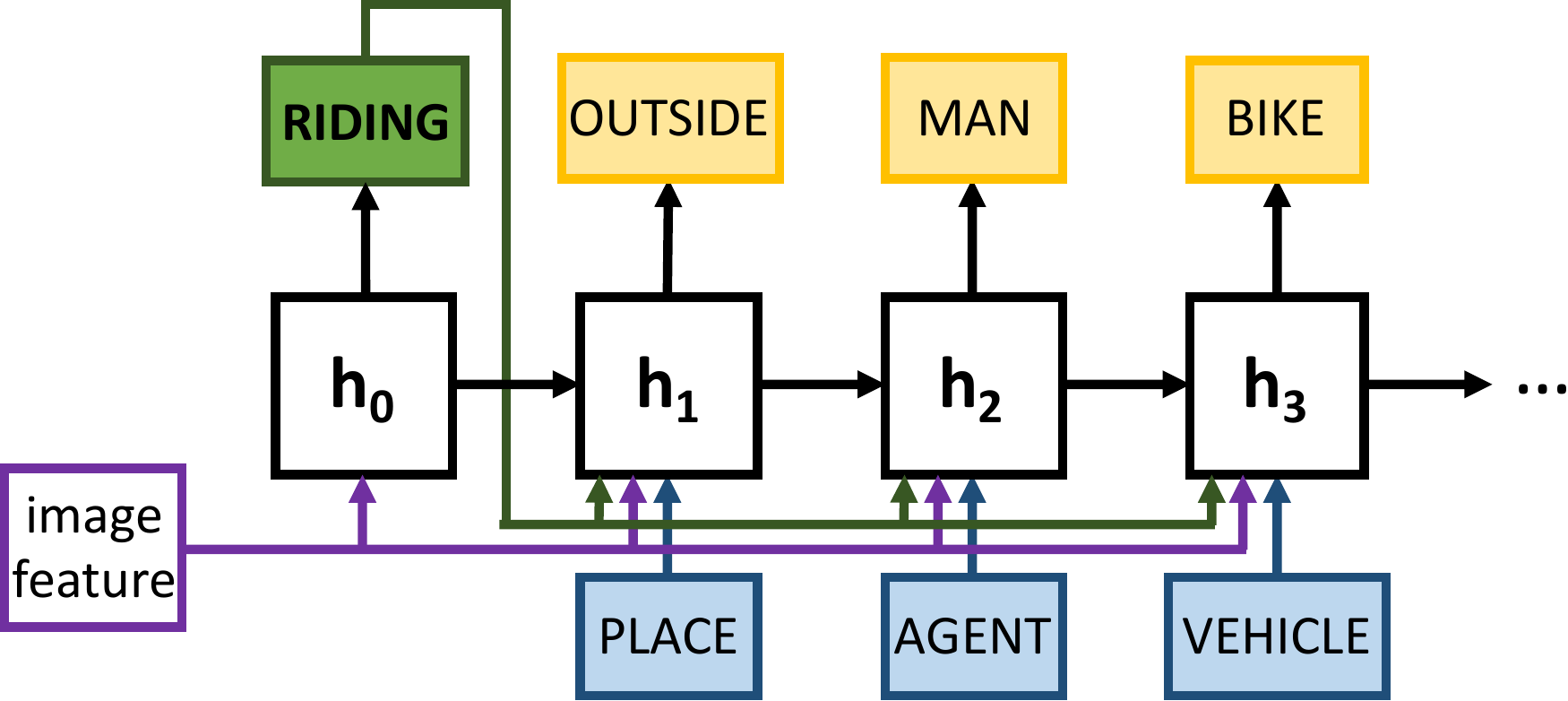}
\caption{The architecture of chain RNN for verb \texttt{riding}.
The time-steps at which different roles are predicted needs to be decided manually, and has an influence on the performance.}
\label{fig:chain_rnn}
\end{figure}

Each image $i$ in the \emph{imSitu} dataset comes with three sets of annotations (from three annotators) for the nouns.
During training, we accumulate the cross-entropy loss at verb and noun nodes for every annotation as
\vspace{-0.2cm}
\begin{equation}
\label{eq:train_loss}
L = \sum_i \sum_{j=1}^{3} \big( y_v \log(p_v) + \frac{1}{|E_f|} \sum_e y_{e:n} \log(p_{e:n}) \big) \, ,
\vspace{-0.2cm}
\end{equation}
where $y_v$ and $y_{e:n}$ correspond to the ground-truth verb for image $i$ and the ground-truth noun for role $e$ of the image, respectively.
Different to the Soft-OR loss in~\cite{yatskar2016imsitu}, we encourage the model to predict all three annotations for each image.
We use back-propagation through time (BPTT)~\cite{werbos1988generalization} to train the model. 

\vspace{-2mm}
\paragraph{Inference.}
At test time, our approach first predicts the verb $\hat{v} = \arg\,\max_v p_v$ to choose a corresponding frame $f$ and obtain the set of associated roles $E_f$.
We then propagate information among role nodes and choose the highest scoring noun $\hat{n}_e = \arg\,\max_n p_{e:n}$ for each role.
Thus our predicted situation is
\vspace{-0.2cm}
\begin{equation}
\hat{S} = (\hat{v}, \{(e, \hat{n}_e): e \in E_f\}) \, .
\end{equation}
To reduce reliance on the quality of verb prediction, we explore beam search over verbs as discussed in Experiments.

\subsection{Simpler Graph Architectures}
\label{subsec:models:rnn}

An alternative to model dependencies between nodes is to use recurrent neural networks (RNN).
Here, situation recognition can be considered as a sequential prediction problem of choosing the verb and corresponding noun-role pairs.
The hidden state of the RNN carries information across the verb and noun-role pairs, and the input at each time-step dictates what the RNN should predict.

\vspace{-0.2cm}
\paragraph{Chain RNN.}
An unrolled RNN can be seen as a special case of a GGNN, where nodes form a chain with directed edges between them.
However, there are a few notable differences, wherein the nodes receive information only once from their (left) neighbor.
In addition, the nodes do not perform $T$ steps of propagation among each other and predict output immediately after the information arrives.

In the standard chain structure of a RNN, we need to manually specify the order of the verb and roles.
As the choice of the verb dictates the set of roles in the frame, we predict the verb at the first time step.
We observe that the \emph{imSitu} dataset and any verb-frame in general, commonly consist of \texttt{place} and \texttt{agent}-like roles (\eg~semantic role \texttt{teacher} can be considered as the \texttt{agent} for the verb \texttt{teaching}).
We thus predict \texttt{place} and \texttt{agent} roles as the second and third roles in the chain
\footnote{Predicting \texttt{place} requires a more global view of the image compared to \texttt{agent}.
Changing the order to \texttt{verb $\rightarrow$ agent $\rightarrow$ place $\rightarrow \ldots$} results in $1.9\%$ drop of performance.}.
We make all other roles for the frame to follow subsequently in descending order of the number of times they occur across all verb-frames.
Fig.~\ref{fig:chain_rnn} shows an example of such a model.

For a fair comparison to the fully connected roles GGNN, we employ the GRU update in our RNN.
The input to the hidden states matches node initialization (Eqs.~\ref{eq:init-verb} and \ref{eq:init-noun}).
We follow the same scheme for predicting the output (linear layer with softmax), and train the model with the same cross-entropy loss.

\input{tables/ablative_study}

\begin{figure}[t]
\centering
\includegraphics[width=1\linewidth]{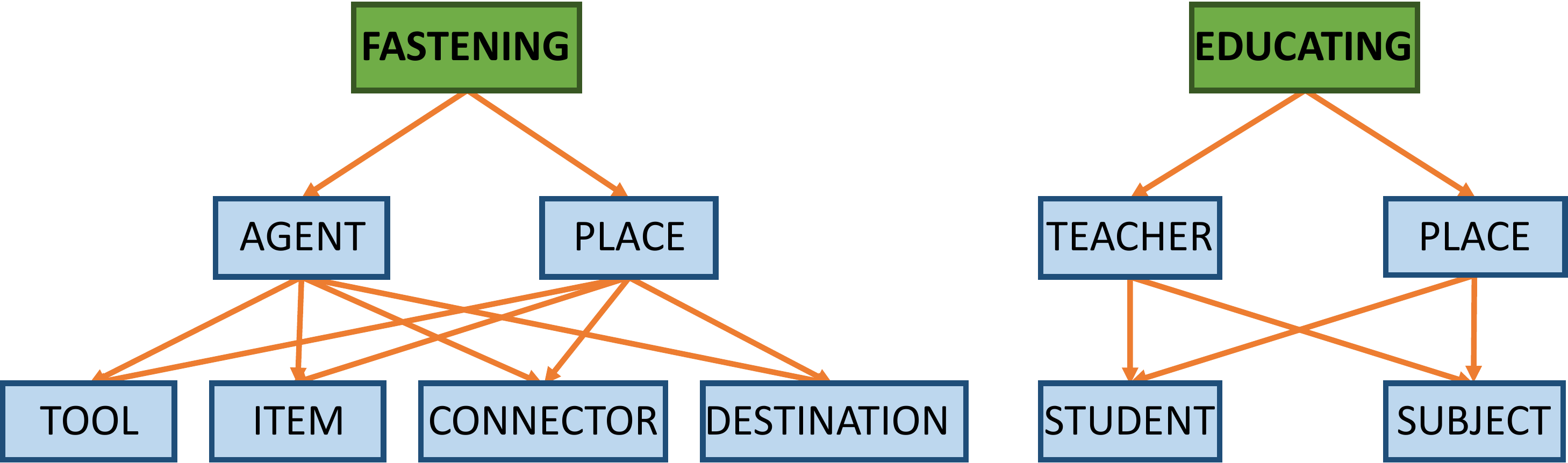}
\caption{The architecture of tree-structured RNN.
Like the Chain RNN, verb prediction is at the root of the tree, and semantic roles \texttt{agent}-like and \texttt{place} are parents of all other roles.}
\label{fig:tree_rnn}
\vspace{-3mm}
\end{figure}

\vspace{-0.2cm}
\paragraph{Tree-structured RNN.}
As mentioned above, the \texttt{place} and \texttt{agent} semantic roles occur more frequently.
We propose a structure where they have a larger chance to influence prediction of other roles.
In particular, we create a tree-structured RNN~\cite{tai2015improved} where the hidden states first predict the verb, followed by \texttt{agent} and \texttt{place}, and all other roles.
Fig.~\ref{fig:tree_rnn} shows examples of resulting structures.

The tree-structured RNN can be deemed as a special case of GGNN, where nodes have the following directed edges:
{\small
\begin{equation}
\mathcal{B} = \{(a_v, a') : a' \in \mathcal{Z}\} \cup \{(a', a) : a' \in \mathcal{Z}, a \in E_f \backslash \mathcal{Z}\} \, ,
\end{equation}}
where $\mathcal{Z} = \{\texttt{agent}, \texttt{place}\}$, and $E_f \backslash \mathcal{Z}$ represents all roles in that frame other than \texttt{agent} and \texttt{place}.
Similar to the chain RNN, we use GRU update and follow the same learning and inference procedures.


%% file: tables/ablative_study.tex

\begin{table*}[t]
\centering
\tabcolsep=0.16cm
{\small
\begin{tabular}{c | l | c c c | c c c | c c | c}
\toprule

& \multirow{2}{*}{Method} & \multicolumn{3}{|c}{top-1 predicted verb} & \multicolumn{3}{|c}{top-5 predicted verbs} & \multicolumn{2}{|c|}{ground truth verbs} & \\
& & verb & value & value-all & verb & value & value-all & value & value-all & mean \\

\midrule
1 & Unaries                                        & 36.32 & 23.74 & 13.86 & 61.51 & 38.57 & 20.76 & 58.32 & 27.57 & 35.08 \\
2 & Unaries, BS=10                                 & 36.39 & 23.74 & 14.01 & 61.65 & 38.64 & 20.96 & 58.32 & 27.57 & 35.16 \\
\midrule
3 & FC Graph, $T$=1                                & 36.25 & 25.99 & 17.02 & 61.60 & 42.91 & 26.44 & 64.87 & 35.52 & 38.83 \\
4 & FC Graph, $T$=2                                & 36.43 & 26.08 & 17.22 & 61.52 & 42.86 & 26.38 & 65.31 & 35.86 & 38.96 \\
5 & FC Graph, $T$=4                                & 36.46 & 26.26 & 17.48 & 61.42 & 43.06 & 26.74 & 65.73 & 36.43 & 39.19 \\
6 & FC Graph, $T$=4, BS=10                         & 36.70 & 26.52 & 17.70 & 61.63 & 43.34 & 27.09 & 65.73 & 36.43 & 39.39 \\
7 & FC Graph, $T$=4, BS=10, vOH                    & \textbf{36.93} & \textbf{27.52} & \textbf{19.15} & \textit{61.80} & \textit{45.23} & \textit{29.98} & \textit{68.89} & \textit{41.07} & \textit{41.32} \\
8 & FC Graph, $T$=4, BS=10, vOH, $g$=ReLU          & 36.26 & 27.22 & \textit{19.10} & \textbf{62.14} & \textbf{45.59} & \textbf{30.32} & \textbf{69.35} & \textbf{41.71} & \textbf{41.46} \\
9 & FC Graph, $T$=4, BS=10, vOH, Soft-OR           & \textit{36.75} & \textit{27.33} & 18.94 & 61.69 & 44.91 & 29.41 & 68.29 & 40.25 & 40.95 \\

\bottomrule
\end{tabular}
}
\vspace{0.1cm}
\caption{Situation prediction results on the development set.
We compare several variants of our fully-connected roles model to show the improvements achieved at every step.
$T$ refers to the number of \textbf{time-steps} of propagation in the fully connected roles GGNN (FC Graph).
\textbf{BS=10} indicates the use of beam-search with beam-width of 10.
\textbf{vOH} (verb, one-hot) is included when the embedding of the predicted verb is used to initialize the hidden state of the role nodes.
\textbf{$g$=ReLU} refers to the non-linear function used after initialization. All other rows use $g$=$\tanh(\cdot)$.
Finally, \textbf{Soft-OR} refers to the loss function used in~\cite{yatskar2016imsitu}.
Best performance is in \textbf{bold} and second-best is \textit{italicized}.}
\label{tab:ablative_study}
\vspace{-0.2cm}
\end{table*}

%% file: main_sections/evaluation.tex

\section{Evaluation}
\label{sec:evaluation}

We evaluate our methods on the \textit{imSitu} dataset~\cite{yatskar2016imsitu} and use the standard splits with 75k, 25k, and 25k images for the \emph{train}, \emph{development}, and \emph{test} subsets, respectively.
Each image in \emph{imSitu} is associated with one verb and three annotations for the role-noun pairs.

We follow~\cite{yatskar2016commonly} and report three metrics:
(i) \emph{verb}: the verb prediction performance;
(ii) \emph{value}: the semantic verb-role-value tuple prediction performance that is considered to be correct if it matches any of the three ground truth annotators; and
(iii) \emph{value-all}: the performance when the \emph{entire} situation is correct and all the semantic verb-role-value pairs match at least one ground truth annotation.

\subsection{Implementation Details}
\label{subsec:eval:details}
\vspace{-0.1cm}
\paragraph{Image Representations.}
We adopt two pre-trained VGG-16 CNNs~\cite{simonyan2015verydeep} for extracting image features by removing the last fully-connected and softmax layers, and fine-tuning all weights.
The first CNN ($\phi_v(i)$) is trained to predict verbs, and second CNN ($\phi_n(i)$) predicts the top $K$ most frequent nouns ($K = 2000$ cover about 95\% of nouns) in the dataset.

\vspace{-0.15in}
\paragraph{Unaries.}
Creating a graph with no edges, or equivalently with $T=0$ steps of propagation corresponds to using the initialized features to perform prediction.
We refer to this approach as \emph{Unaries}, which will be used as the simplest baseline to showcase the benefit of modeling dependencies between the roles.

\vspace{-0.15in}
\paragraph{Learning.}
We implement the proposed models in Torch~\cite{torch}.
The network is trained using RMSProp~\cite{hinton2012lecture} with mini-batches of 256 samples.
We choose the hidden state dimension $D = 1024$, and train image ($W_{iv}, W_{in}$), verb ($W_v$) and role ($W_e$) embeddings.
The image features are extracted before training the GGNN or RNN models.

The initial learning rate is $10^{-3}$ and starts to decay after 10 epochs by a factor of $0.85$.
We use dropout with a probability of $0.5$ on the output prediction layer (\cf~Eqs.~\ref{eq:pred-verb} and \ref{eq:pred-noun}) and clip the gradients to range $(-1, 1)$.

\vspace{-0.15in}
\paragraph{Mapping \texttt{agent} Roles.}
The \emph{imSitu} dataset~\cite{yatskar2016imsitu} has situations for 504 verbs.
Among them, we notice that 19 verbs do not have the semantic role \texttt{agent} but instead with roles of similar meaning (\eg~verb \texttt{educating} has role \texttt{teacher}).
We map these alternative roles to \texttt{agent} when determining their position in the RNN architecture.
Such a mapping is not used for the fully connected GGNN model.

\vspace{-0.15in}
\paragraph{Variable Number of Roles.}
A verb has a maximum of 6 roles associated with it.
We implement our proposed model with fixed-size graphs involving 7 nodes.
To deal with verbs with less than 6 roles, we zero the hidden states at each time-step of propagation, making them not receive or send any information.

\input{tables/graph_structures}
\input{tables/sota_comparison}

\subsection{Results}
\label{subsec:eval:results}

We first present a quantitative analysis comparing different variants of our proposed model.
We then evaluate the performance of different architectures, and compare results with state-of-the-art approaches.

\vspace{-0.1in}
\paragraph{Ablative Analysis}
A detailed study of the GGNN model with fully connected roles (referred to as \emph{FC Graph}) is shown in Table~\ref{tab:ablative_study}.
An important hyper-parameter for the GGNN model is the number of propagation steps $T$.
We found that the performance increases by a small amount when increasing $T$, and saturates soon (in rows 3, 4, and 5).
We believe that this is due to the use of a fully-connected graph, and all nodes sharing most of the information at the first-step propagation.
Nevertheless, the propagation is important, as revealed in the comparison between \emph{Unaries} ($T=0$) from row 1 and $T=1$ in row 3.
We obtain a mean improvement of 3.8\% in all metrics.

During test we have the option of using beam search, where we hold $B$ best verb predictions and compute the role-noun predictions for each of the corresponding graphs (frames). 
Finally, we select the top prediction using the highest log-probability across all $B$ options.
We use a beam width of $B=10$ in our experiments, which yields small improvement.
Rows 1 and 2 of Table~\ref{tab:ablative_study} show the improvement using beam search on a graph without propagation.
Rows 5 and 6 show the benefit after multiple steps of propagation.

Rows 6 and 7 of Table~\ref{tab:ablative_study} demonstrate the impact of using embeddings of the predicted verb (vOH) to initialize the role nodes' hidden states in Eq.~(\ref{eq:init-noun}).
Notable improvement is obtained when using the ground-truth verb (3-4\%). The \emph{value-all} for the top-1 predicted verb increases from 17.70\% to 19.15\%.
We also tested different non-linear functions for initialization, \ie, $\tanh$ (row 7) or ReLU (row 8), however, the impact is almost negligible.
We thus use $\tanh$ for all experiments.

Finally, comparing rows 7 and 9 of Table~\ref{tab:ablative_study} reveals that our loss function to predict all annotations in Eq.~(\ref{eq:train_loss}) performs slightly better than the Soft-OR loss that aims to fit at least one of the annotations~\cite{yatskar2016imsitu}.

\begin{figure}[t]
\vspace{-1mm}
\centering
\includegraphics[width=1.0\linewidth]{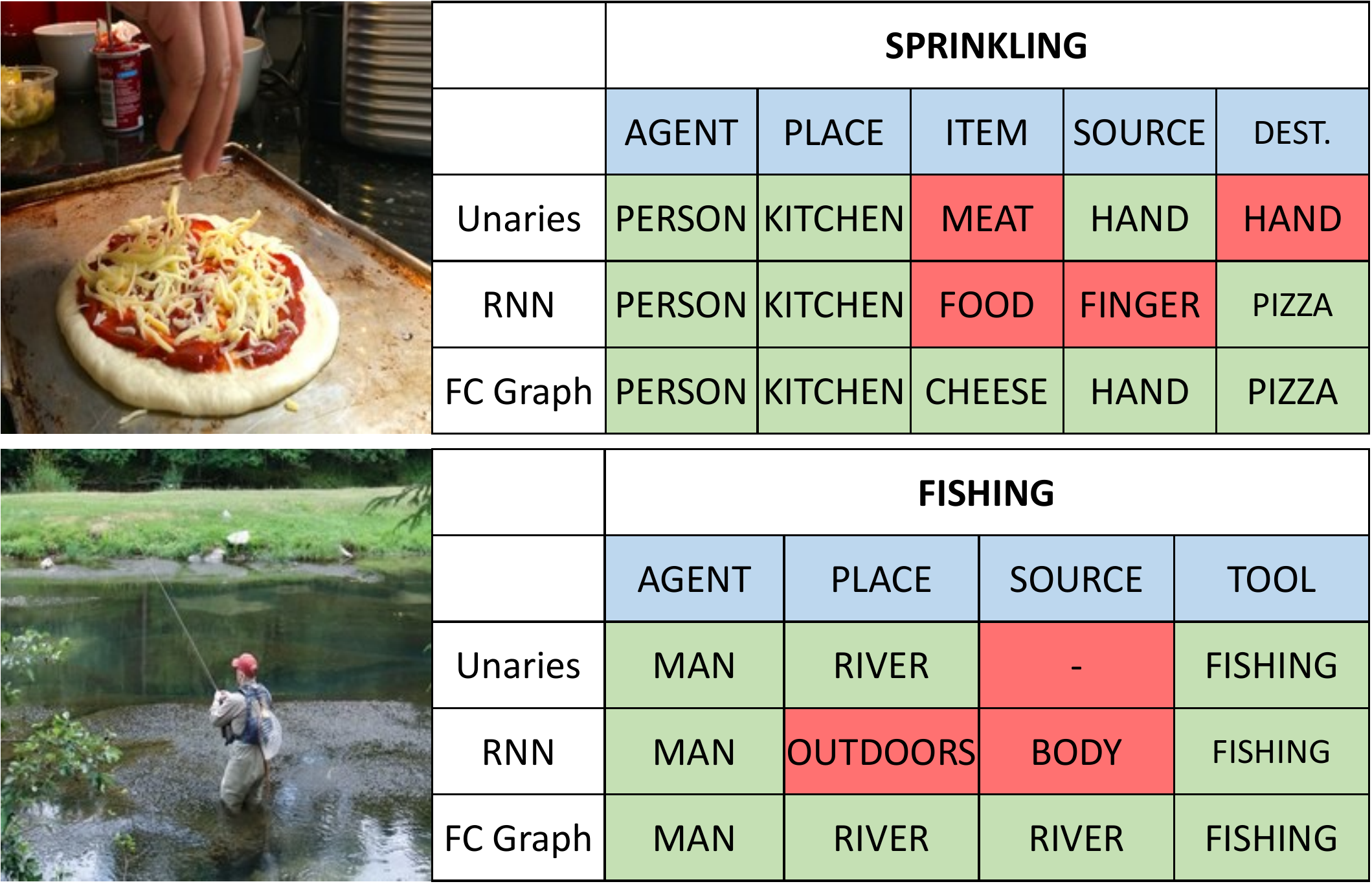}
\caption{Example images with their predictions listed from all methods.
Roles are marked with a blue background, and predicted nouns are in green boxes when correct, and red when wrong.
Using the \emph{FC Graph} corrects mistakes made by the \emph{Unaries} or \emph{Chain RNN} prediction models.}
\label{fig:structures_improve}
\vspace{-4mm}
\end{figure}

\vspace{-2mm}
\paragraph{Baseline RNNs.}
Table~\ref{tab:graph_structures} summarizes the results with different structures on the dev set.
As expected, \emph{Unaries} perform consistently worse than models with information propagation between nodes on the \emph{value} and \emph{value-all} metrics.
The \emph{Tree-structured RNN} provides a $2\%$ boost in \emph{value-all} for top-1 predicted verb, while the \emph{Chain RNN} provides a $3.9\%$ improvement.
Owing to the better connectivity between the roles in a \emph{Chain RNN} (especially \texttt{place} and \texttt{agent}), we observe better performance compared to the \emph{Tree-structured RNN}.
Note that as the RNNs are trained jointly to predict both verbs and nouns, and as the noun gradients dominate, the verb prediction takes a hit.

\vspace{-2mm}
\paragraph{Different Graph Structures.}
We can also use \emph{chain} or \emph{tree-structured} graphs in GGNN.
Along with the FC graph in row 6 of Table~\ref{tab:graph_structures}, rows 4 and 5 present the results for different GGNN structures.
They show that connecting roles with each other is critical and sharing information helps.
Interestingly, the Chain GGNN needs more propagation steps ($T$=8), as it takes time for the left-most and right-most nodes to share information.
Smaller values of $T$ are possible when nodes are well-connected as in Tree-structured ($T$=6) or FC Graph ($T$=4).
Fig.~\ref{fig:structures_improve} presents prediction from all models for two images.
The \emph{FC Graph} is able to reason about associating \texttt{cheese} and \texttt{pizza} rather than \texttt{sprinkling} \texttt{meat} or \texttt{food} on it.

\vspace{-2mm}
\paragraph{Comparison with State-of-the-art.}
We compare the performance of our models against state-of-the-art on both the dev and test sets in Table~\ref{tab:sota_comparison}.
Our CNN predicts the verb well.
Beam search leads to even better performance (2-4\% higher) in verb prediction.
We note that \emph{Tensor Composition + DataAug} actually uses more data to train models.
Nevertheless, we achieve the best performance on all metrics when using the top-1 predicted verb.

Another advantage of our model is in improvement for the \emph{value-all} metric.
It yields +8\% when using the ground-truth verb, +6\% with top-5 predicted verbs, and +4.5\% with top-1 predicted verb, compared with the baseline without data augmentation.
Interestingly, even with data augmentation, we outperform~\cite{yatskar2016commonly} by 3-4\% in \emph{value-all} for top-1 predicted verb.
This property attributes to information sharing between role nodes, which helps in correcting errors and better predicts frames.
Note that \emph{value-all} is an important metric to measure a full understanding of the image. 
Models with higher \emph{value-all}  will likely lead to better captioning or question-answering results.

\subsection{Further Discussion}
\label{subsec:eval:discuss}

We delve deeper into our model and discuss why the \emph{FC Graph} outperforms baselines.

\begin{figure}[t]
\centering
\includegraphics[width=1.0\linewidth]{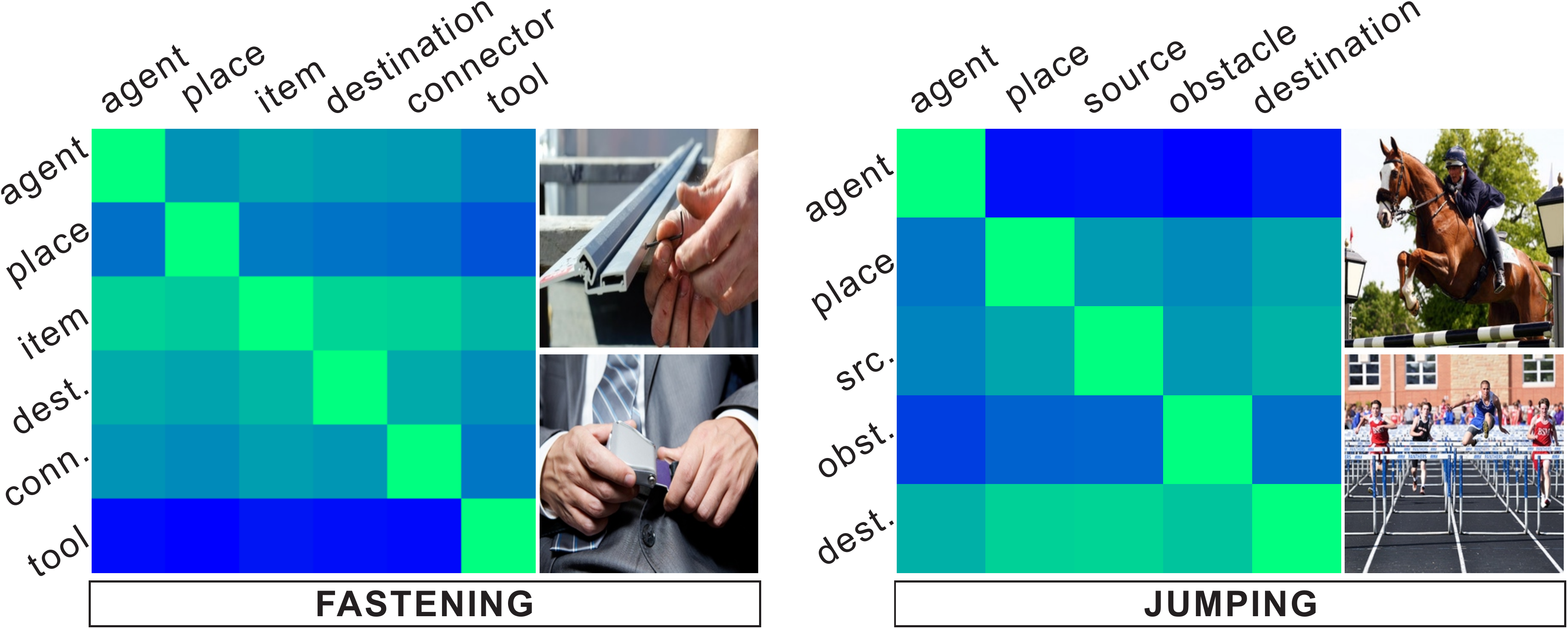}
\caption{We present the ``amount'' of information that is propagated between roles for two verbs along with sample images.
Blue corresponds to high, and green to zero.
Each element of the matrix corresponds to the norm of the incoming message from different roles (normalized column sum to 1).
\textbf{Left:} verb \texttt{fastening} needs to pay attention to the \texttt{tool} used. 
\textbf{Right:} important components to describe \texttt{jumping} are the \texttt{agent} and \texttt{obstacles} along the path.
}
\label{fig:learned_structures}
\vspace{-0.2cm}
\end{figure}

\begin{figure}[t]
\centering
\includegraphics[width=0.9\linewidth]{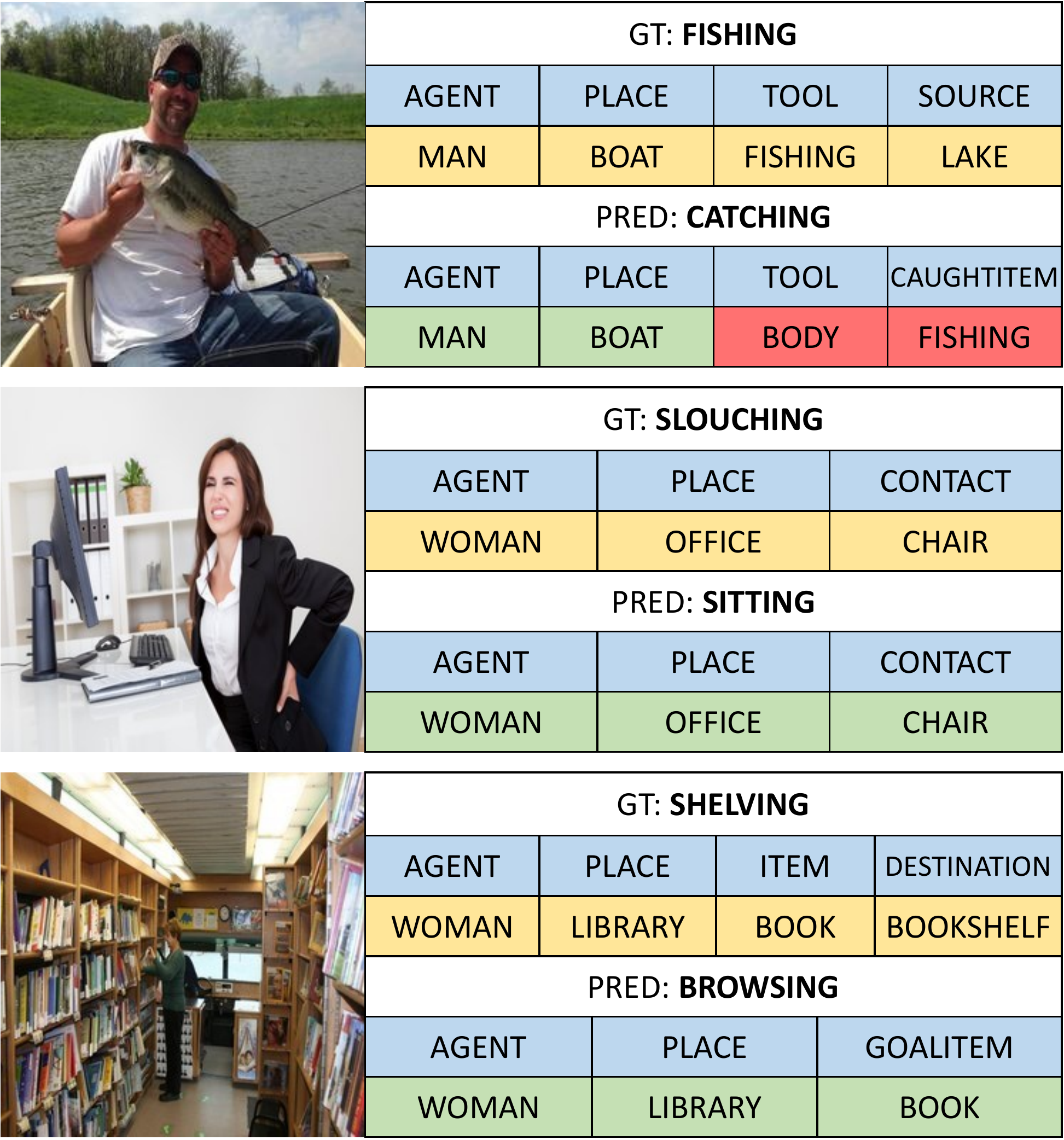}
\caption{Images with ground-truth and top-1 predictions from the development set.
Roles are marked with blue background.
Ground-truth (GT) nouns are in yellow and predicted (PRED) nouns with green when correct, or red when wrong.
Although the predicted verb is different from the ground-truth, it is very plausible.
Some of the verbs refer to the same frame (\eg~\texttt{sitting} and \texttt{slouching}), and contain the same set of roles, which our model is able to correctly infer.
}
\label{fig:wrong_verb}
\vspace{-0.4cm}
\end{figure}

\begin{figure*}[t]
\centering
\includegraphics[width=0.96\linewidth]{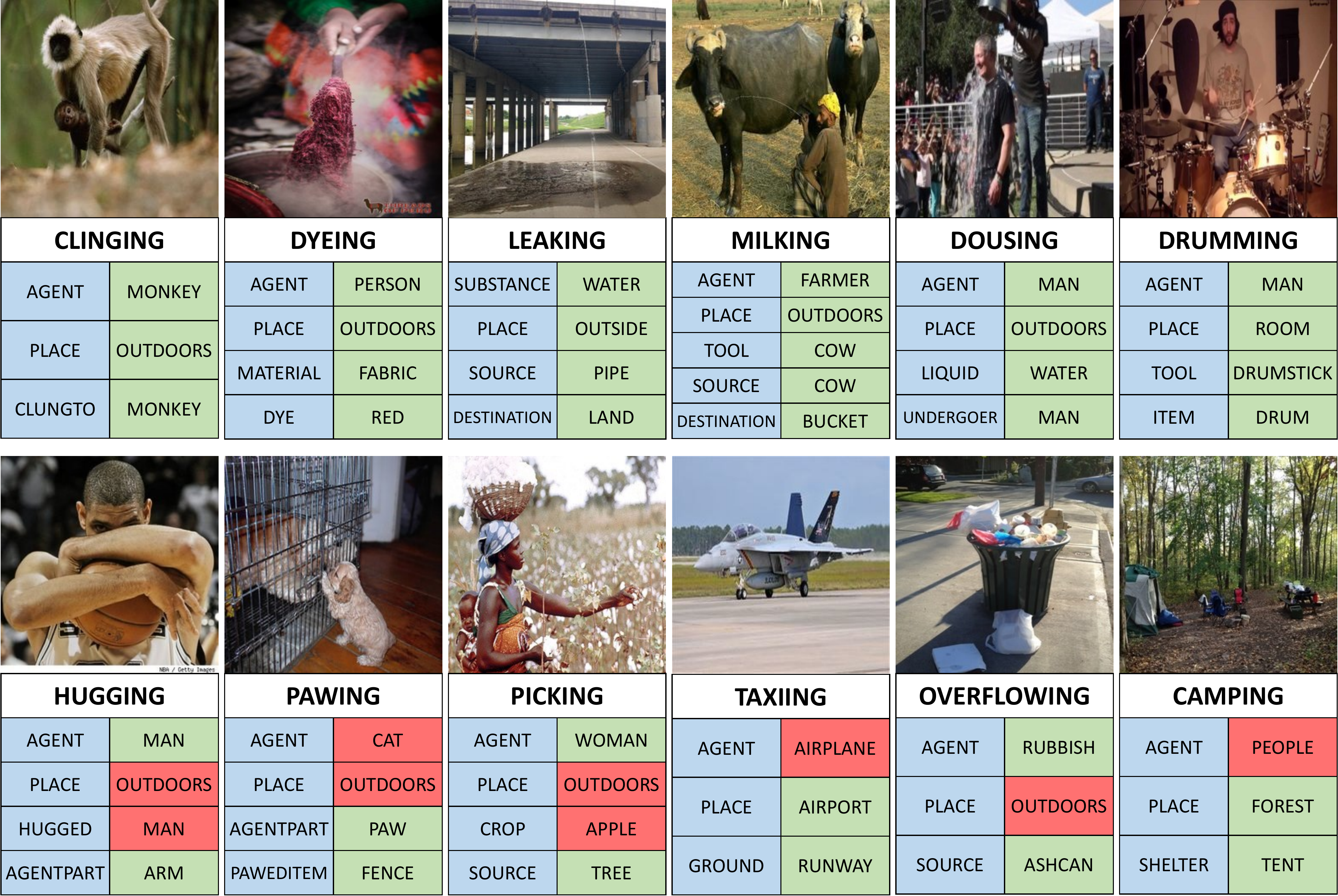}
\caption{Images with top-1 predictions from the development set.
For all samples, the predicted verb is correct, shown below the image in bold.
Roles are marked with a blue background, and predicted nouns are in green when correct, and red when wrong.
\textbf{Top row}: We are able to correctly predict the situation (verb and all role-noun pairs) for all samples.
\textbf{Bottom row}: The first three samples contain errors in prediction (\eg~the \texttt{agent} for the verb \texttt{pawing} is clearly a \texttt{dog}).
However, the latter three samples are in fact correct predictions that are not found in the ground-truth annotations (\eg~\texttt{people} are in fact \texttt{camping} in the \texttt{forest}).}
\label{fig:last_page}
\vspace{-0.2cm}
\end{figure*}

\vspace{-0.1in}
\paragraph{Learned Structure.}
A key emphasis of this model is on information propagation between roles.
In Fig.~\ref{fig:learned_structures}, we present the norms of the propagation matrices.
Each element in the matrix $P(a', a)$ is the norm of the incoming message from role $a'$ to $a$ averaged across all images (in dev set) at the first time-step, i.e., $\|x_{(a',a)}^{t=1}\|$ regarding Eq.~(\ref{eq:hidden}).
In this example, \texttt{tool} is important for the verb \texttt{fastening} and influences all other roles, while \texttt{agent} and \texttt{obstacle} influence roles in \texttt{jumping}.


\vspace{-0.1in}
\paragraph{Wrong Verb Predictions.}
We present a few examples of top scoring results where the verb prediction is wrong in Fig.~\ref{fig:wrong_verb}.
Note that in fact these predicted verbs are plausible options for the given images.
The metric \emph{value} treats them as wrong, and yet we can correctly predict the role-noun pairs.
One example is the middle one of \texttt{slouching} vs. \texttt{sitting}.
Fig.~\ref{fig:wrong_verb} (bottom) shows that choosing a different verb might lead to the selection of different roles (\texttt{goalitem} vs. \texttt{item, destination}).
Nevertheless, predicting \texttt{book} for \texttt{browsing} is a good choice.

\vspace{-0.1in}
\paragraph{Predictions with Correct Verb.}
Fig.~\ref{fig:last_page} shows several examples of prediction obtained by FC Graph, where the predicted verb matches the ground-truth one.
The top row corresponds to samples where the metric \emph{value-all} scores correctly as all role-noun pairs are correct.
Note that the roles are closely related (\eg.~\texttt{(agent, clungto)} and \texttt{(material, dye)}) and help each other choose the correct nouns.
In the bottom row, we show some failure cases in predicting role-noun pairs.
First, the model favors predicting \texttt{place} as \texttt{outdoor} (a majority of place is \texttt{outdoor} in the training set).
Second, for the sample with verb \texttt{picking}, we predict the \texttt{crop} as \texttt{apple}, which appears 79 times in the dataset compared with \texttt{cotton} that appears 14 times.
Providing more training samples (\eg~\cite{yatskar2016commonly}) could help remedy such issues.

In the latter three samples of the bottom row, although the model makes reasonable predictions, they do not match the ground-truth.
For example, the ground-truth annotation for the verb \texttt{taxiing} is \texttt{agent:jet} and for the verb \text{camping} is \texttt{agent:persons}.
Therefore, even though each image comes with three annotations, synonymous nouns and verbs make the task still challenging.

%% file: tables/graph_structures.tex

\begin{table*}[t]
\centering
\tabcolsep=0.16cm
{\small
\begin{tabular}{c | l | c c c | c c c | c c | c}
\toprule

&  & \multicolumn{3}{|c}{top-1 predicted verb} & \multicolumn{3}{|c}{top-5 predicted verbs} & \multicolumn{2}{|c|}{ground truth verbs} & \\
&  & verb & value & value-all & verb & value & value-all & value & value-all & mean \\

\midrule
1 &    Unaries                                           & \textit{36.39} & 23.74 & 14.01 & \textit{61.65} & 38.64 & 20.96 & 58.32 & 27.57 & 35.16 \\
2 &    Chain RNN                                         & 34.62 & \textit{24.67} & \textit{17.94} & 61.09 & \textit{41.67} & \textit{27.80} & \textit{62.58} & \textit{36.57} & \textit{38.36} \\
3 &    Tree-structured RNN                               & 34.62 & 24.24 & 16.04 & 58.86 & 39.15 & 23.65 & 60.44 & 30.91 & 35.98 \\
\midrule
4 &    Chain GGNN, $T$=8                                 & 36.63 & 27.27 & 19.03 & \textbf{61.88} & 44.97 & 29.44 & 68.20 & 40.21 & 40.95 \\
5 &    Tree-structured GGNN, $T$=6                       & \textit{36.78} & \textit{27.48} & \textbf{19.54} & 61.75 & \textit{45.12} & \textbf{30.11} & \textit{68.54} & \textit{41.01} & \textit{41.29} \\
6 &    Fully-connected GGNN, $T$=4                       & \textbf{36.93} & \textbf{27.52} & \textit{19.15} & \textit{61.80} & \textbf{45.23} & \textit{29.98} & \textbf{68.89} & \textbf{41.07} & \textbf{41.32} \\

\bottomrule
\end{tabular}
}
\vspace{0.1cm}
\caption{Situtation prediction results on the development set for models with different graph structures.
All models use beam search, predicted verb embedding, and $g = \tanh(\cdot)$.
Best performance is highlighted in \textbf{bold}, and second-best in each table section is \textit{italicized}.
}
\vspace{0cm}
\label{tab:graph_structures}
\end{table*}

%% file: tables/sota_comparison.tex

\begin{table*}[t]
\centering
\tabcolsep=0.16cm
{\small
\begin{tabular}{ c | l | c c c | c c c | c c | c}
\toprule

  & & \multicolumn{3}{|c}{top-1 predicted verb} & \multicolumn{3}{|c}{top-5 predicted verbs} & \multicolumn{2}{|c|}{ground truth verbs} & \\
  & & verb & value & value-all & verb & value & value-all & value & value-all & mean \\

\midrule
\multirow{5}{*}{\rot{dev}}
    & CNN+CRF~\cite{yatskar2016imsitu}                         & 32.25 & 24.56 & 14.28 & 58.64 & 42.68 & 22.75 & 65.90 & 29.50 & 36.32 \\
    & Tensor Composition~\cite{yatskar2016commonly}            & 32.91 & 25.39 & 14.87 & 59.92 & 44.50 & 24.04 & \textit{69.39} & 33.17 & 38.02 \\
    & Tensor Composition + DataAug~\cite{yatskar2016commonly}  & 34.20 & \textit{26.56} & 15.61 & \textbf{62.21} & \textbf{46.72} & 25.66 & \textbf{70.80} & 34.82 & \textit{39.57} \\
    & Chain RNN                                                 & \textit{34.62} & 24.67 & \textit{17.94} & 61.09 & 41.67 & \textit{27.80} & 62.58 & \textit{36.57} & 38.36 \\
    & Fully-connected Graph                                     & \textbf{36.93} & \textbf{27.52} & \textbf{19.15} & \textit{61.80} & \textit{45.23} & \textbf{29.98} & 68.89 & \textbf{41.07} & \textbf{41.32} \\

\midrule
\multirow{5}{*}{\rot{test}}
    & CNN+CRF~\cite{yatskar2016imsitu}                         & 32.34 & 24.64 & 14.19 & 58.88 & 42.76 & 22.55 & 65.66 & 28.96 & 36.25 \\
    & Tensor Composition~\cite{yatskar2016commonly}            & 32.96 & 25.32 & 14.57 & 60.12 & 44.64 & 24.00 & \textit{69.20} & 32.97 & 37.97 \\
    & Tensor Composition + DataAug~\cite{yatskar2016commonly}  & 34.12 & \textit{26.45} & 15.51 & \textbf{62.59} & \textbf{46.88} & 25.46 & \textbf{70.44} & 34.38 & \textit{39.48} \\
    & Chain RNN                                                 & \textit{34.63} & 24.65 & \textit{17.89} & 61.06 & 41.73 & \textit{28.15} & 62.94 & \textit{37.32} & 38.54 \\
    & Fully-connected Graph                                     & \textbf{36.72} & \textbf{27.52} & \textbf{19.25} & \textit{61.90}  & \textit{45.39} & \textbf{29.96} & 69.16 & \textbf{41.36} & \textbf{41.40} \\

\bottomrule
\end{tabular}
}
\vspace{0.1cm}
\caption{We compare situation prediction results on the development and test sets against state-of-the-art models.
Each model was run on the test set only once.
Our model shows significant improvement in the top-1 prediction on all metrics, and performs better than a baseline that uses data augmentation.
The performance improvement on the \emph{value-all} metric is important for applications, such as captioning and QA.
Best performance is highlighted in \textbf{bold}, and second-best is \textit{italicized}.
}
\vspace{-0cm}
\label{tab:sota_comparison}
\end{table*}

%% file: main_sections/conclusion.tex

\section{Conclusion}
\label{sec:conclusion}

We presented an approach for recognizing situations in images that involves predicting the correct verb along with its corresponding frame consisting of role-noun pairs.
Our Graph Neural Network (GNN) approach explicitly models dependencies between verb and roles, allowing nouns to inform each other.
On a benchmark dataset~\emph{imSitu}, we achieved $\sim$4.5\% accuracy improvement on a metric that evaluates correctness of the entire frame (\emph{value-all}).
We presented analysis of our model, demonstrating the need to capture the dependencies between roles, and compared it with RNN models and other related solutions.

%% file: main_sections/appendix.tex

\onecolumn
\newpage
\begin{center}
\textbf{\Large{
Supplementary Material
}}
\end{center}
\vspace{0.5cm}

We present additional analysis and results of our approach in the supplementary material.
First, we analyze the verb prediction performance in Sec.~\ref{sec:verb_pred}.
In Sec.~\ref{sec:tsne_embeddings}, we present t-SNE~\cite{maaten2008tsne} plots to visualize the verb and role embeddings.
We present several examples of the influence of different roles on predicting the \emph{verb-frame} correctly.
This is visualized in Sec.~\ref{sec:prop_matrices} through propagation matrices similar to Fig.~7 of the main paper.
Finally, in Sec.~\ref{sec:predictions} we include several example predictions that our model makes.

\section{Verb Prediction}
\label{sec:verb_pred}
\begin{figure}[ht]
\centering
\includegraphics[width=0.9\linewidth]{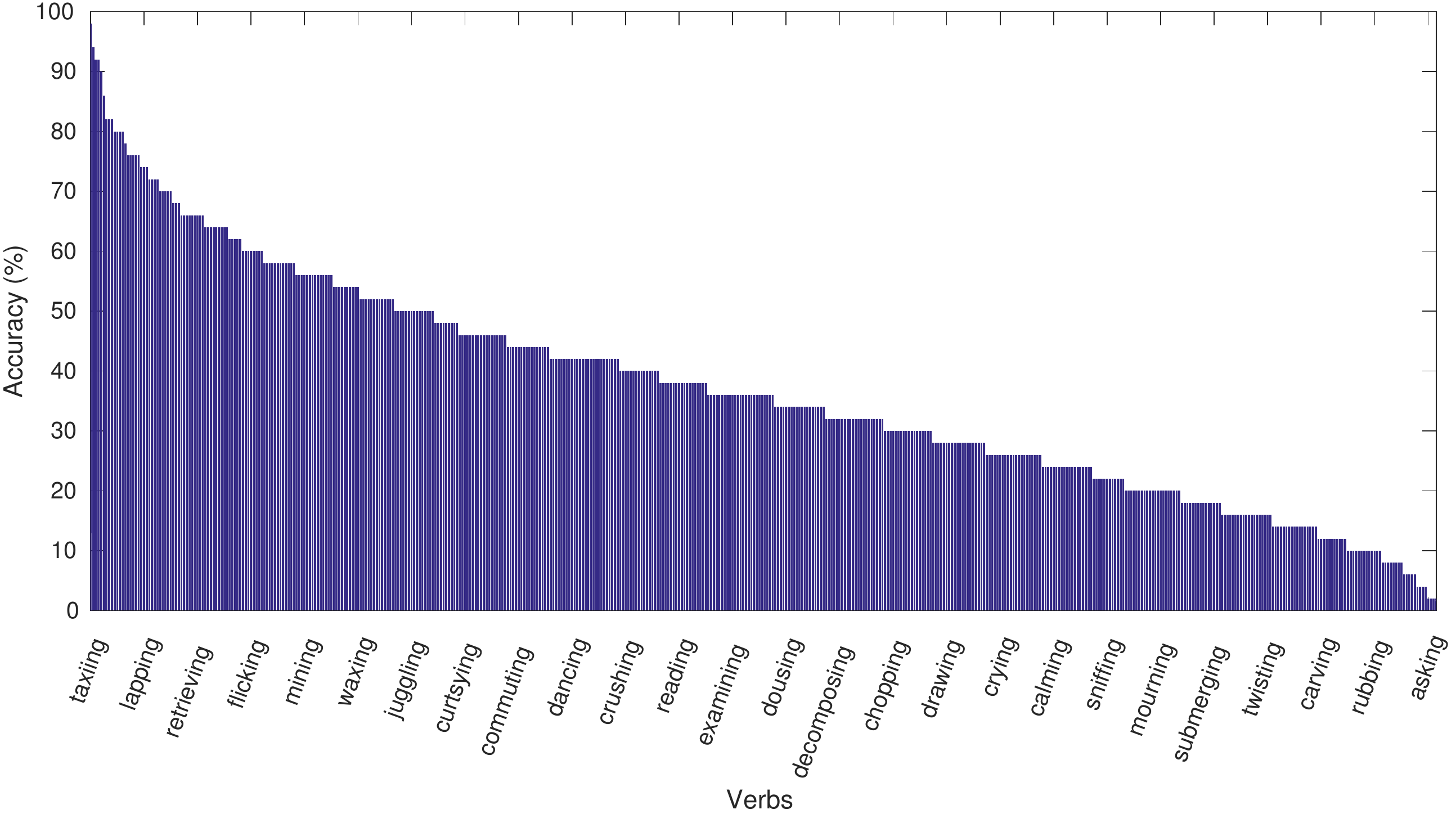}
\caption{Verb prediction accuracy on the development set.
Some verbs such as \texttt{taxiing} typically have a similar image (a plane on the tarmac), while verbs such as \texttt{rubbing} or \texttt{twisting} can have very different corresponding images.}
\label{fig:verb_accuracy}
\end{figure}

We present the verb prediction accuracies for our fully-connected model on the development set in Fig.~\ref{fig:verb_accuracy}.
The random performance is close to 0.2\% (504 verbs).
About 22\% of all verbs are classified correctly over 50\% of the time.
These include \texttt{taxiing, erupting, flossing, microwaving,}~\etc.
On the other hand, verbs such as \texttt{attaching, making, placing} can have very different image representations, and show prediction accuracies of less than 10\%.

Our model helps improve the role-noun predictions by sharing information across all roles.
Nevertheless, if the verb is predicted incorrectly, the whole situation is treated as incorrect.
Thus, verb prediction performance plays a crucial role.

\paragraph{Confusion between similar verbs.}
We analyze the confusion between similar verbs, that according to the metrics, leads to incorrect situation recognition.
In the main paper, Fig.~8 presents a few examples where we are able to correctly predict the roles, but the situation is classified as wrong since the verb is incorrect.

The \emph{imSitu} dataset consists of 504 verbs, and while we do have a complete $504\times504$ confusion matrix, visualizing the results is hard.
As explained in the dataset~\cite{yatskar2016imsitu}, the verb frames were obtained using FrameNet.
We notice that the 504 verbs from the \emph{imSitu} dataset are grouped into 161 FrameNet verbs~\cite{framenet}.
For example, several verbs such as \texttt{walking, climbing, skipping, prowling} and 26 others are clustered together to the FrameNet verb: \texttt{self\_motion}.
The clusters need not be large, and 73 of 161 clusters consist of just one verb.

\begin{figure}[ht]
\centering
\includegraphics[width=0.32\linewidth]{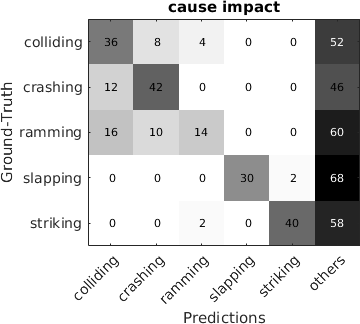} \,
\includegraphics[width=0.32\linewidth]{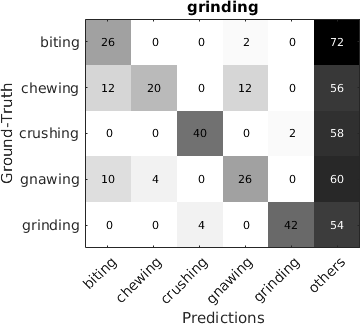} \,
\includegraphics[width=0.32\linewidth]{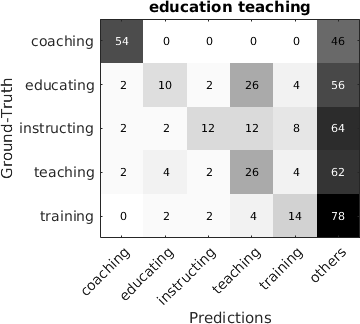} \\\vspace{0.2cm}
\includegraphics[width=0.32\linewidth]{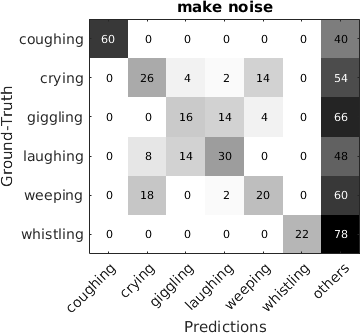} \,
\includegraphics[width=0.32\linewidth]{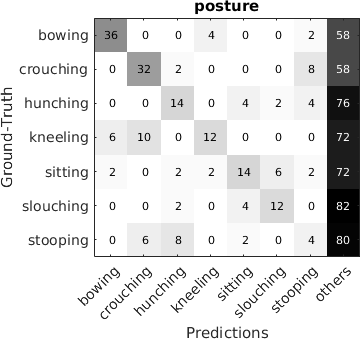} \,
\includegraphics[width=0.32\linewidth]{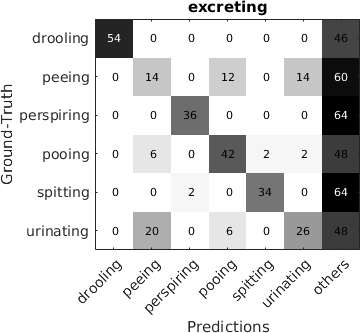}
\caption{Confusion matrices for verb prediction.
Each row indicates the expected ground-truth, and the columns are predictions (each row sums to 100\%).
As it is not possible to show all 504 verbs, we pick verb clusters based on their FrameNet labels (shown in the title).
Confusion between remaining verbs not in the cluster is grouped in the last column as \texttt{others}.
The examples show significant confusion between verbs which are hard to differentiate visually: \texttt{colliding-crashing-ramming}, or \texttt{crying-giggling-laughing-weeping}.}
\label{fig:confusion_matrices}
\end{figure}

We use this as a clustering, and present several confusion matrices for verb clusters in Fig.~\ref{fig:confusion_matrices}.
All verb predictions that do not belong to the cluster are grouped as \texttt{others}.
While, the others column does collect most of the predictions, there is significant confusion between similar verbs.

\section{Verb and Role Embeddings}
\label{sec:tsne_embeddings}

We initialize the hidden states of our role nodes (\cf~Eq.~2 of the main paper) with
\begin{equation}
h_{a_e}^0 = g(W_{in} \phi_n(i) \odot W_e e \odot W_v \hat{v}) \, ,
\end{equation}
where, $W_v$ and $W_e$ are verb and role embeddings respectively, and $e \in \mathbb{R}^{190}$ and $\hat{v} \in \mathbb{R}^{504}$ are one-hot vectors representing the noun for a specific role, and the predicted verb.
$\phi_n(i)$ is the image representation using the noun-prediction CNN.
Note that both verbs and roles are embedded to a $\mathbb{R}^{1024}$ space.

\begin{figure}[ht]
\centering
\includegraphics[width=\linewidth]{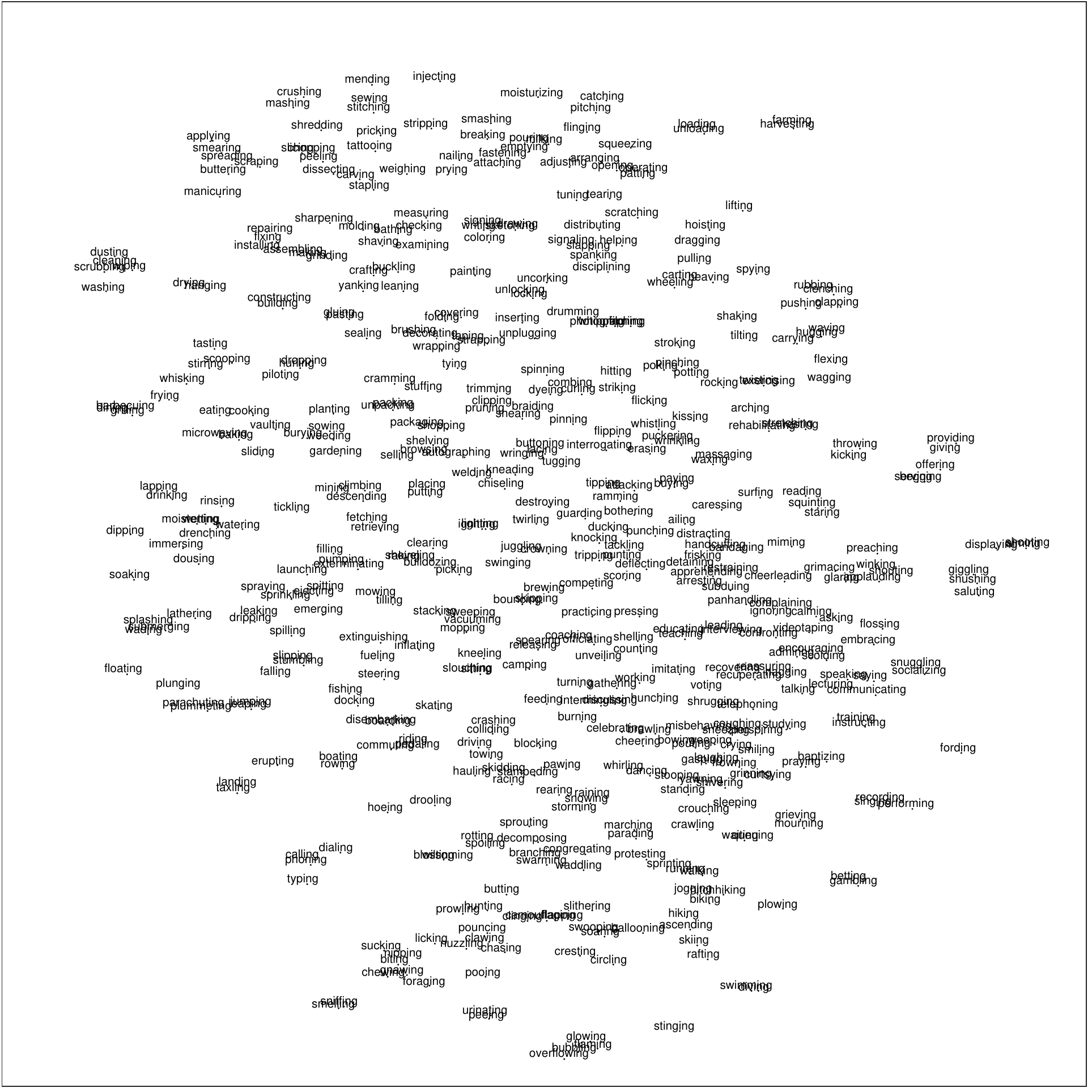}
\caption{2D t-SNE representation of the all the learned verb embeddings.
While the number of labels is quite large, it is still possible to see small clusters of verbs forming at the periphery of the figure.
\textbf{top}: farming-harvesting, pouring-emptying-milking, slicing-chopping-peeling.
\textbf{top-right}: carting-wheeling-heaving, pinching-poking.
\textbf{right}: providing-giving, offering-begging-serving, reading-squinting-staring.
\textbf{bottom-right}: betting-gambling, grieving-mourning, baptizing-praying.
\textbf{bottom}: glowing-flaming, bubbling-overflowing, sniffing-smelling.
\textbf{bottom-left}: landing-taxiing, dialing-calling-phoning-typing, boating-rowing.
\textbf{left}: drinking-lapping, microwaving-baking, mining-climbing-descending.
\textbf{top-left}: dusting-scrubbing-cleaning-wiping, drying-hanging, repairing-fixing-installing.
}
\label{fig:tsne_allverb}
\end{figure}

\begin{figure}[ht]
\centering
\includegraphics[width=\linewidth]{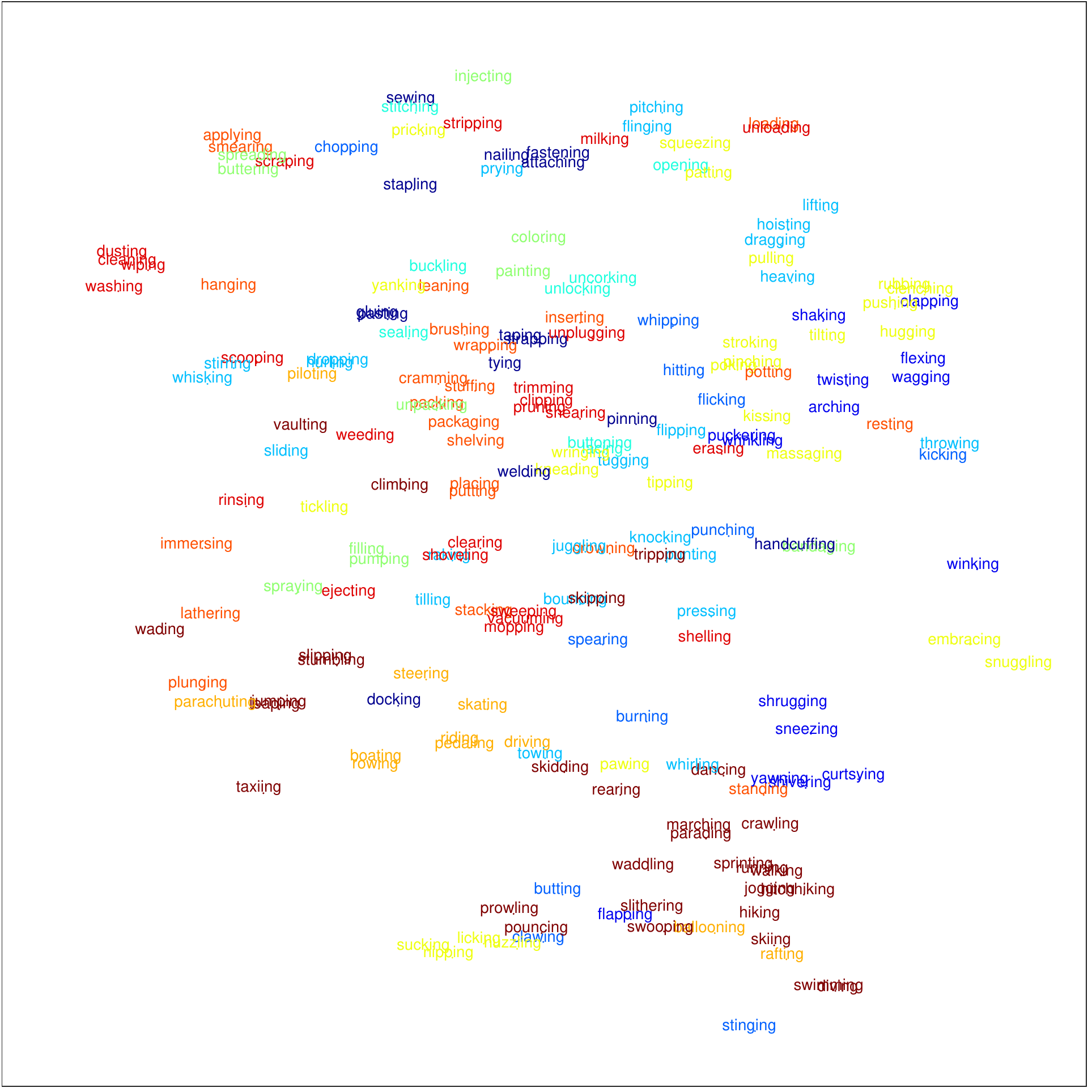}
\caption{2D t-SNE representation of the learned verb embeddings of the verbs belonging to 11 largest clusters (using FrameNet verb clustering).
The clusters are: \emph{attaching, body\_movement, cause\_harm, cause\_motion, closure, filling, manipulation, operate\_vehicle, placing, removing, self\_motion}.
Each cluster is assigned a unique color from the jet colormap.
Our model is even able to learn to embed similar verbs across these FrameNet groupings.
For example, it brings together
\texttt{whirling} (FrameNet: cause\_motion) and \texttt{dancing} (FN: self\_motion);
\texttt{raking} (FN: cause\_motion) and \texttt{shoveling} (FN: removing);
\texttt{packing} (FN: placing) and \texttt{unpacking} (FN: filling);
\texttt{throwing} (FN: cause\_motion) and \texttt{kicking} (FN: cause\_harm); and many others.
}
\label{fig:tsne_selectverb}
\end{figure}

\paragraph{Verbs.}
The dataset consists of 504 verbs.
We first show a plot depicting all verbs in Fig.~\ref{fig:tsne_allverb}.
Owing to the number of verbs, this is quite hard to see, nevertheless, we can still observe clusters of similar verbs (\eg~\texttt{dusting-cleaning- scrubbing-wiping}, \texttt{recording-singing-performing},~\etc.).

Additionally, we use the verb clustering afforded by the FrameNet verb associations, and select a set of 196 verbs from the 11 largest clusters (cluster size $\ge 8$).
We present their embeddings in Fig.~\ref{fig:tsne_selectverb}.
The learned embeddings not only discover the clustering, but are also able to associate across clusters.
For example, (in the top-left corner), \texttt{applying} and \texttt{smearing} belong to the \texttt{Placing} FrameNet verb, while \texttt{spreading} and \texttt{buttering} correspond to \texttt{Filling} in FrameNet.
Nevertheless, our model is able to learn that these verbs may have similar context (\eg~buttering bread), and brings their representations close.

\begin{figure}[ht]
\centering
\includegraphics[width=0.8\linewidth]{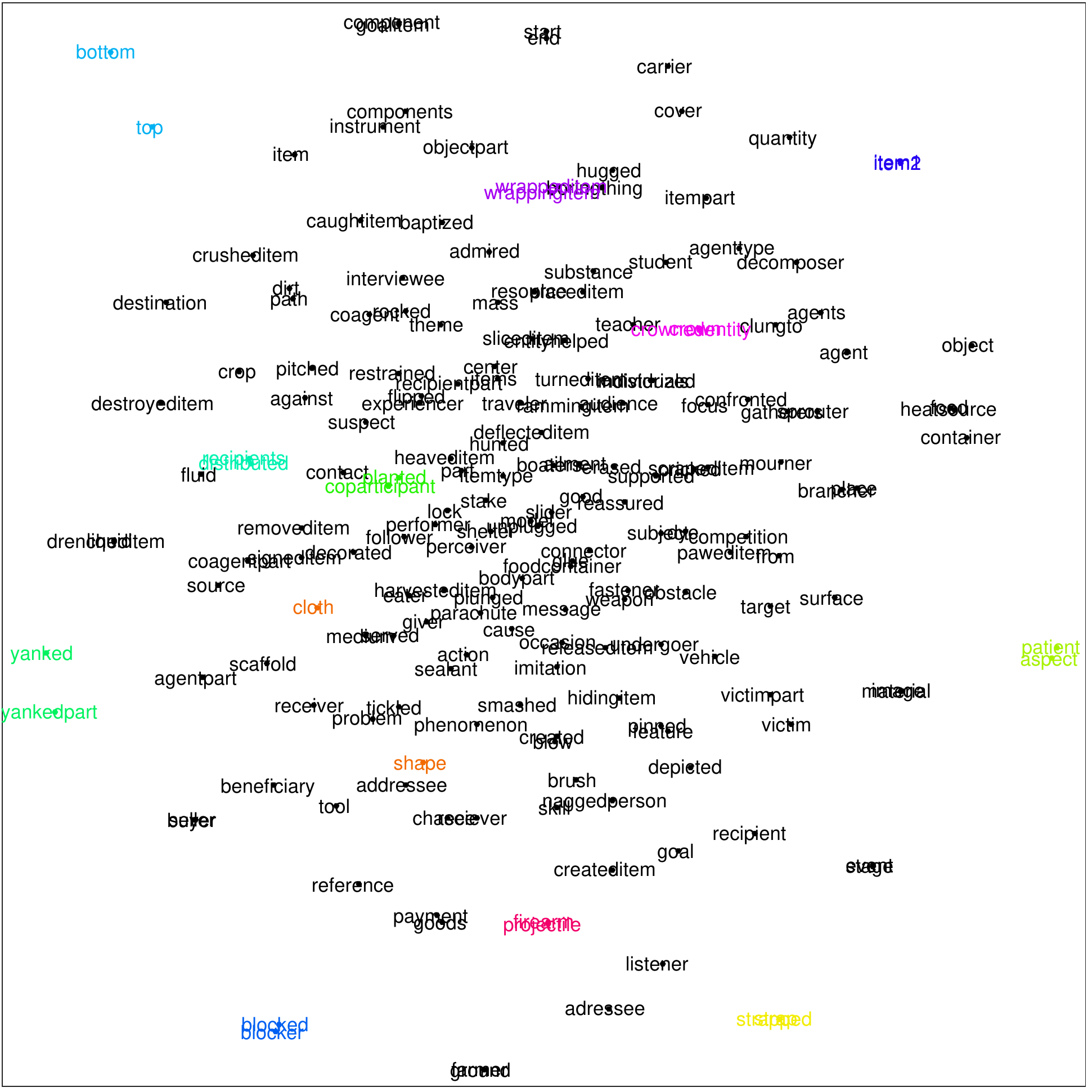}
\caption{2D t-SNE representation of the learned role embeddings.
Note how semantic roles capturing similar themes are brought together.
For example, \texttt{blocked-blocker}, or \texttt{recipients-distributed}, or \texttt{payment-goods}.
Additionally, related semantic roles that apply across verbs are also brought together.
For example, \texttt{components-instrument-object-part}, or \texttt{liquid- drencheditem}, \texttt{foodcontainer-glue-connector}.
As most roles do not present a natural clustering, we are unable to color all roles, and they are shown in black.
Colored roles are associated with one unique verb.}
\label{fig:tsne_role}
\end{figure}

\paragraph{Roles.}
The dataset comes with 190 roles, however, 139 of them are unique to one verb.
For example, the roles \texttt{top} and \texttt{bottom} appear only once, in the frame for the verb \texttt{stacking}.
Similarly, roles \texttt{shape} and \texttt{cloth} appear only when the verb is \texttt{folding}.
We present two-dimensional t-SNE~\cite{maaten2008tsne} representations of the learned role embeddings in Fig.~\ref{fig:tsne_role}.
We associate same colors with role pairs that are associated with only one verb (there are only 12 such pairs, accounting for 24 of 190 roles).
All other roles are shown in black.
In the Fig.~\ref{fig:tsne_role}, we see that the strongly related pairs that are unique to one verb (and colored) are very close to each other.
Additionally, other semantic roles that are related, \eg~\texttt{food}, \texttt{heatsource}, \texttt{container} (right side of figure) are also close together.

\input{supp_sections/promatrix_results_arxiv}

\vspace{1cm}
\section{Prediction Results}
\label{sec:predictions}
\begin{figure}[ht]
\centering
\includegraphics[width=0.9\linewidth]{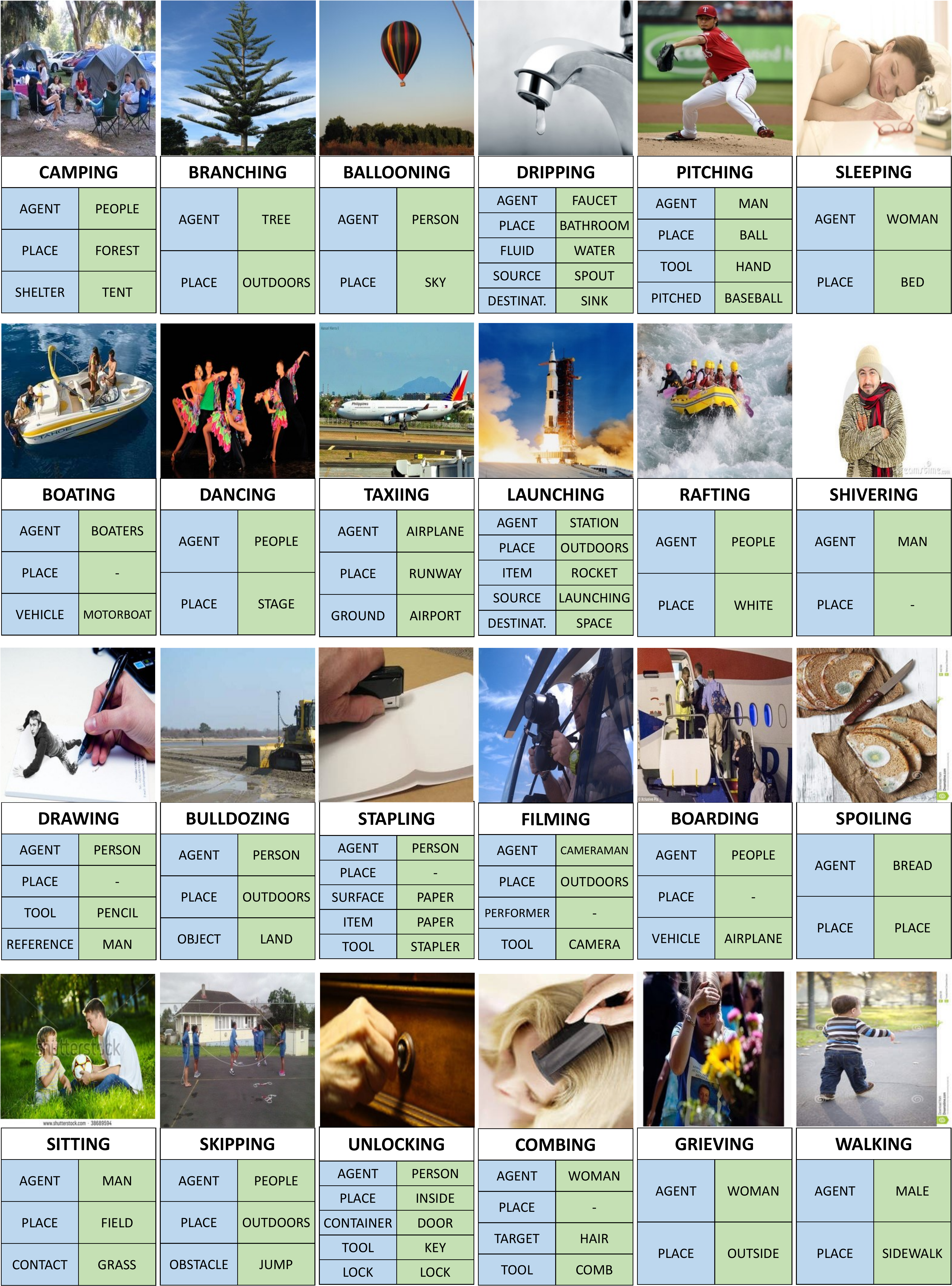}
\caption{Images with top-1 predictions from the development set.
For all samples, the predicted verb is correct, and is shown below the image in bold.
Roles are marked with a blue background, and predicted nouns with green when correct, and red when wrong.
We are able to correctly predict the situation (verb and all role-noun pairs) for all example images shown here.}
\label{fig:supp_examples1}
\end{figure}

\begin{figure}[ht]
\centering
\includegraphics[width=\linewidth]{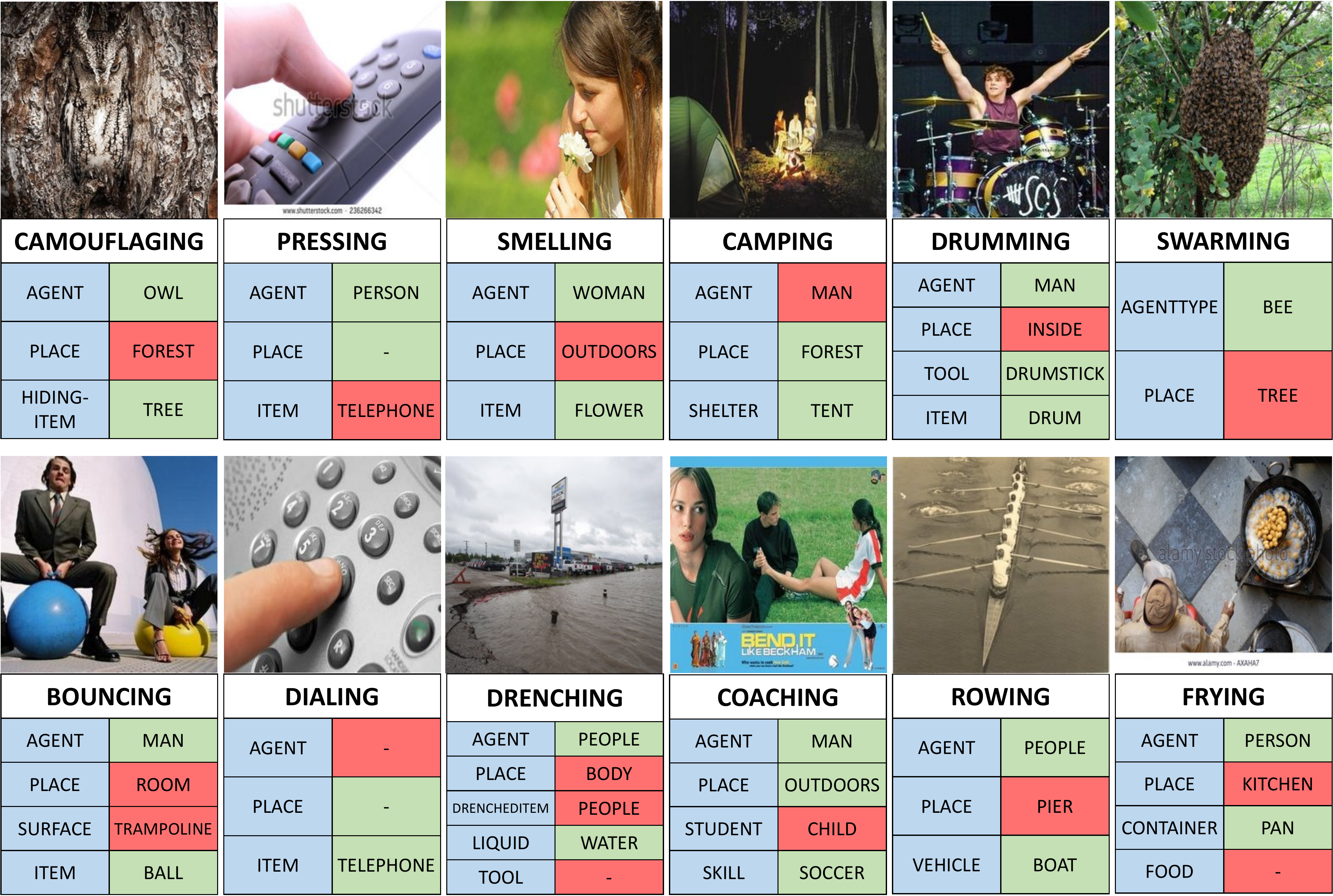}
\caption{Images with top-1 predictions from the development set.
For all samples, the predicted verb is correct, and is shown below the image in bold.
Roles are marked with a blue background, and predicted nouns with green when correct, and red when wrong.
We show examples with genuine errors in prediction (\eg~the \texttt{telephone} for the verb \texttt{pressing} is clearly a \texttt{remote control}).
However, some examples are marked wrong due to the lack of matching ground-truth annotations (\eg~the woman \texttt{smelling} the flower is \texttt{outdoors} (GT: field)).}
\label{fig:supp_examples2}
\end{figure}

\begin{figure}[ht]
\centering
\includegraphics[width=\linewidth]{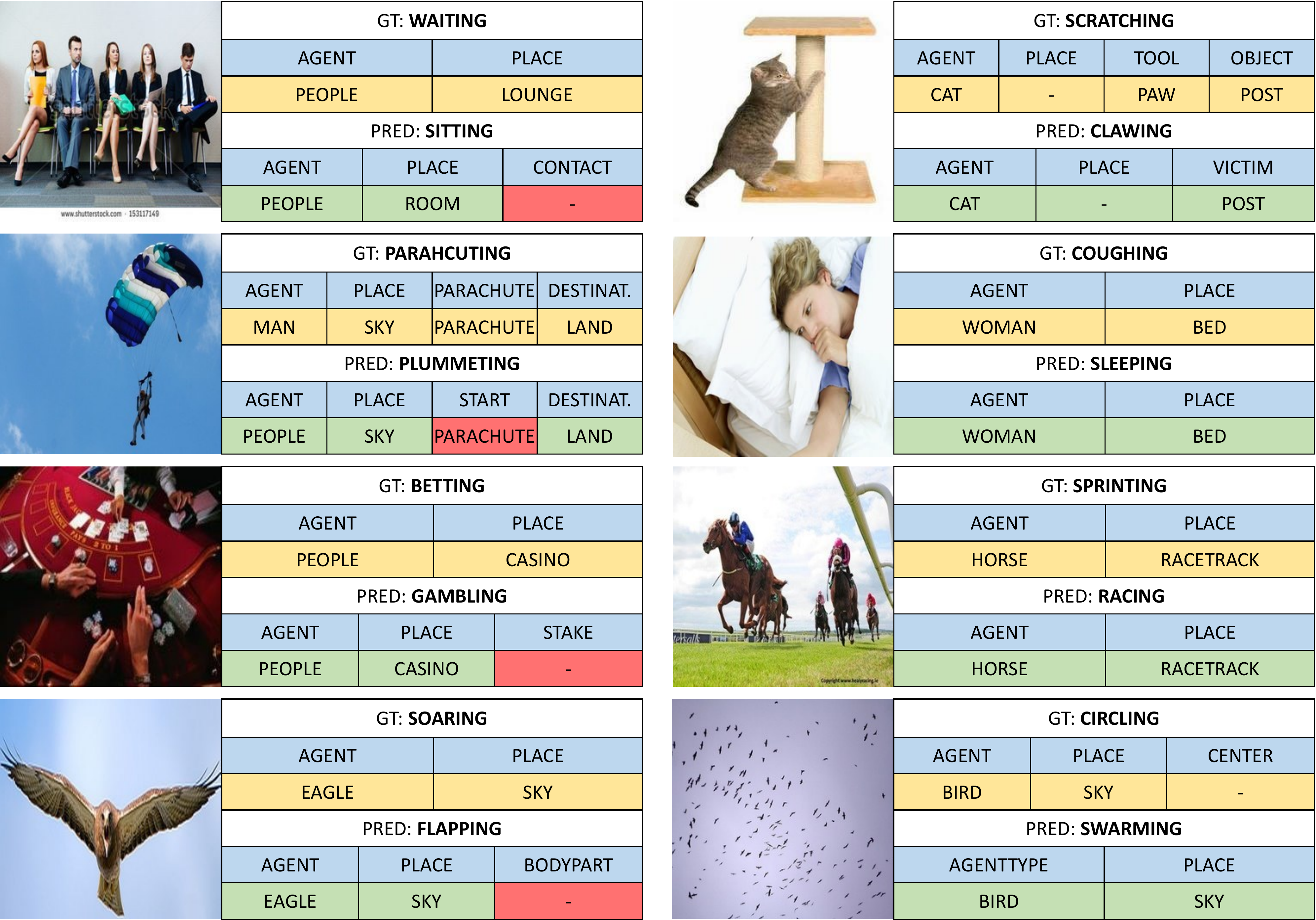}
\caption{Images with ground-truth and top-1 predictions from the development set.
Roles are marked with blue background. Ground-truth (GT) nouns with yellow, and predicted (PRED) nouns with green when correct, or red when wrong.
Although the predicted verb is different from the ground-truth, it is very plausible.
Some of the verbs refer to the same frame (\eg~\texttt{sprinting}, \texttt{racing}), and contain the same set of roles, which our model is able to correctly infer.}
\label{fig:supp_wrongverbs}
\end{figure}

We round up the supplementary material with several more example predictions from our model.
Fig.~\ref{fig:supp_examples1} shows predictions that are completely correct.
Fig.~\ref{fig:supp_examples2} shows examples where we are able to predict the correct verb, but not all the role-noun pairs. Such examples are counted towards the \emph{value} metric, but not \emph{value-all}.
Finally, Fig.~\ref{fig:supp_wrongverbs} shows top-scoring (log-probability) examples where the verb is wrongly predicted, but is mostly plausible (the correct noun predictions are not captured by any metric).
The role-noun pairs here are often correct.

%% file: supp_sections/promatrix_results_arxiv.tex

\newpage
\section{Visualizing the propagation matrices.}
\label{sec:prop_matrices}

We visualize the propagation matrix for 15 more verbs (extending Fig.~7 of the main paper).
Note that, even though we choose the verbs randomly, we see that many verbs do have dominant roles that influence others.
Each row consists of the matrix, and 4 randomly chosen images corresponding to the verb.

Our model propagates information between all roles, and we present the norm of the message sent by each role to the other in the propagation matrix.
The verb and list of roles is displayed at the beginning of each row for simplicity.
The rows and columns of the propagation matrix follow this ordering of roles.

\newpage
\noindent\textbf{Verb:} ADJUSTING \\
\noindent\textbf{Roles:} agent, place, item, feature, tool
\begin{figure}[h]
\centering
\includegraphics [width=0.19\linewidth] {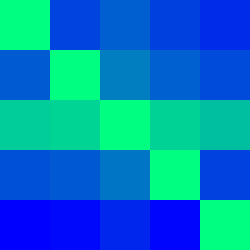}
\includegraphics [width=0.19\linewidth] {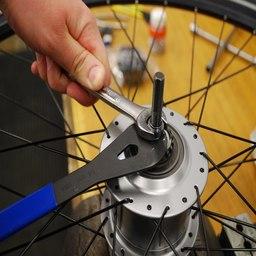}
\includegraphics [width=0.19\linewidth] {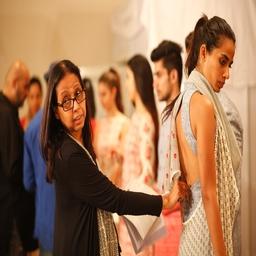}
\includegraphics [width=0.19\linewidth] {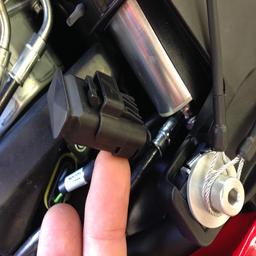}
\includegraphics [width=0.19\linewidth] {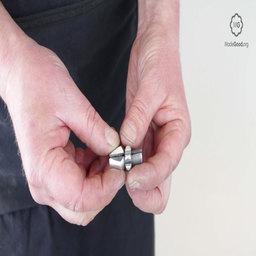}
\end{figure}

\noindent\rule{\textwidth}{1pt}

\noindent\textbf{Verb:} AUTOGRAPHING \\
\noindent\textbf{Roles:} agent, place, item, receiver
\begin{figure}[h]
\centering
\includegraphics [width=0.19\linewidth] {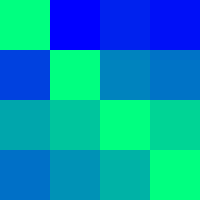}
\includegraphics [width=0.19\linewidth] {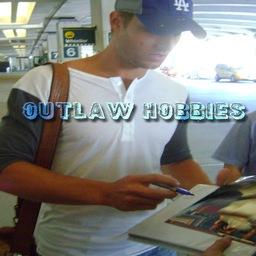}
\includegraphics [width=0.19\linewidth] {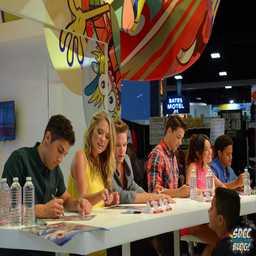}
\includegraphics [width=0.19\linewidth] {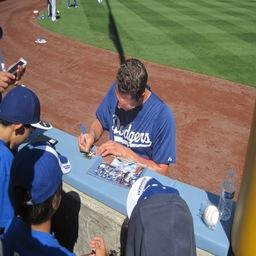}
\includegraphics [width=0.19\linewidth] {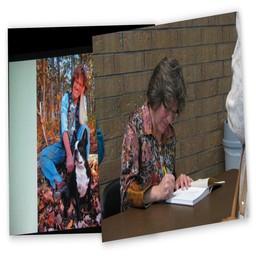}
\end{figure}

\noindent\rule{\textwidth}{1pt}
\noindent\textbf{Verb:} BRUSHING \\
\noindent\textbf{Roles:} agent, place, target, tool, substance
\begin{figure}[h]
\centering
\includegraphics [width=0.19\linewidth] {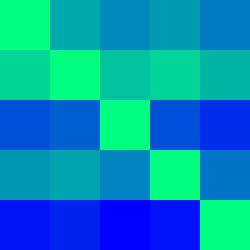}
\includegraphics [width=0.19\linewidth] {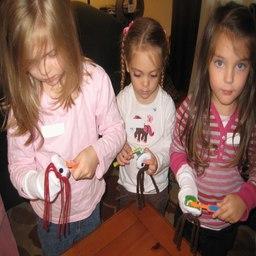}
\includegraphics [width=0.19\linewidth] {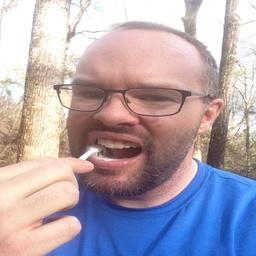}
\includegraphics [width=0.19\linewidth] {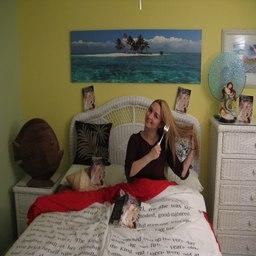}
\includegraphics [width=0.19\linewidth] {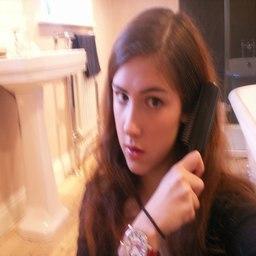}
\end{figure}

\noindent\rule{\textwidth}{1pt}

\noindent\textbf{Verb:} BURNING \\
\noindent\textbf{Roles:} agent, place, target
\begin{figure}[h]
\centering
\includegraphics [width=0.19\linewidth] {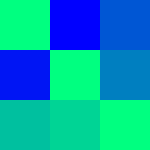}
\includegraphics [width=0.19\linewidth] {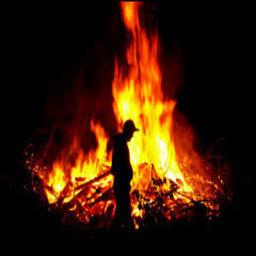}
\includegraphics [width=0.19\linewidth] {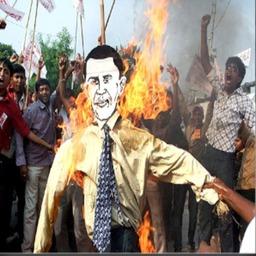}
\includegraphics [width=0.19\linewidth] {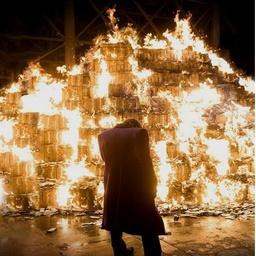}
\includegraphics [width=0.19\linewidth] {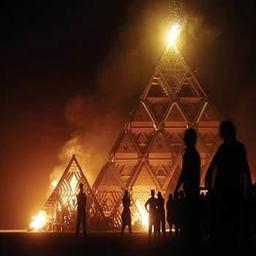}
\end{figure}

\noindent\rule{\textwidth}{1pt}

\newpage
\noindent\textbf{Verb:} CHECKING \\
\noindent\textbf{Roles:} agent, place, patient, aspect, tool
\begin{figure}[h]
\centering
\includegraphics [width=0.19\linewidth] {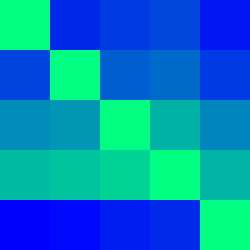}
\includegraphics [width=0.19\linewidth] {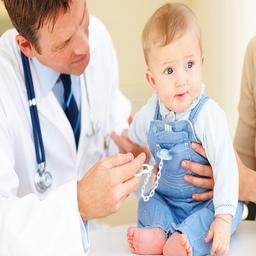}
\includegraphics [width=0.19\linewidth] {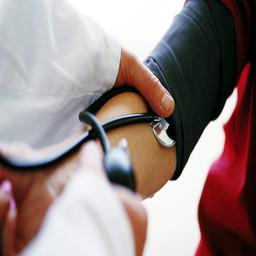}
\includegraphics [width=0.19\linewidth] {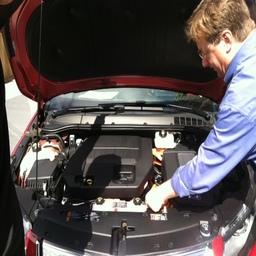}
\includegraphics [width=0.19\linewidth] {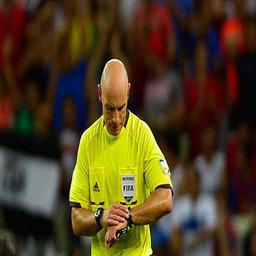}
\end{figure}

\noindent\rule{\textwidth}{1pt}

\noindent\textbf{Verb:} CRAFTING \\
\noindent\textbf{Roles:} agent, place, created, instrument
\begin{figure}[h]
\centering
\includegraphics [width=0.19\linewidth] {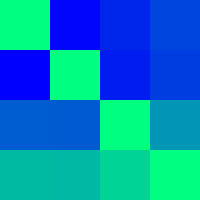}
\includegraphics [width=0.19\linewidth] {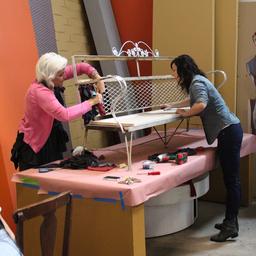}
\includegraphics [width=0.19\linewidth] {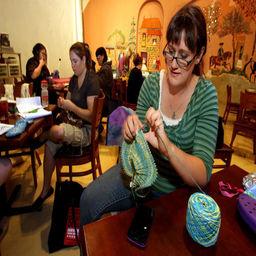}
\includegraphics [width=0.19\linewidth] {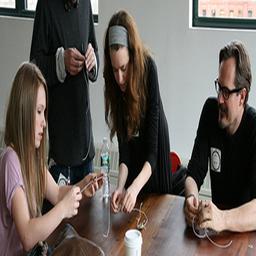}
\includegraphics [width=0.19\linewidth] {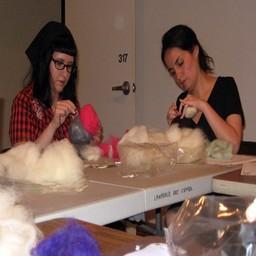}
\end{figure}

\noindent\rule{\textwidth}{1pt}
\noindent\textbf{Verb:} DIPPING \\
\noindent\textbf{Roles:} agent, place, item, substance
\begin{figure}[h]
\centering
\includegraphics [width=0.19\linewidth] {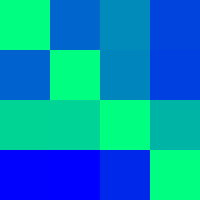}
\includegraphics [width=0.19\linewidth] {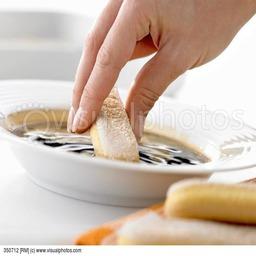}
\includegraphics [width=0.19\linewidth] {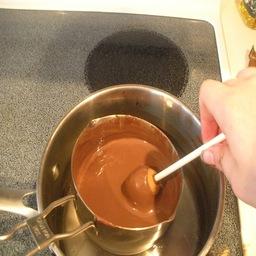}
\includegraphics [width=0.19\linewidth] {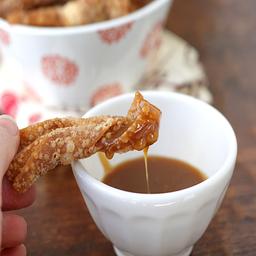}
\includegraphics [width=0.19\linewidth] {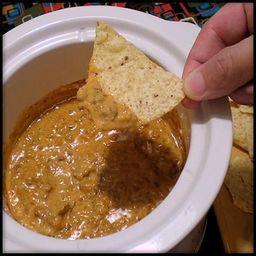}
\end{figure}

\noindent\rule{\textwidth}{1pt}

\noindent\textbf{Verb:} DISTRIBUTING \\
\noindent\textbf{Roles:} agent, place, tool, distributed, recipients
\begin{figure}[h]
\centering
\includegraphics [width=0.19\linewidth] {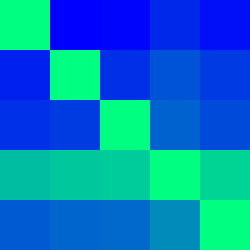}
\includegraphics [width=0.19\linewidth] {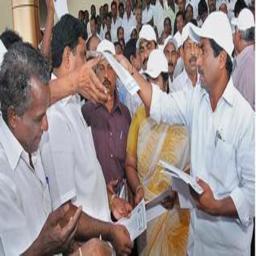}
\includegraphics [width=0.19\linewidth] {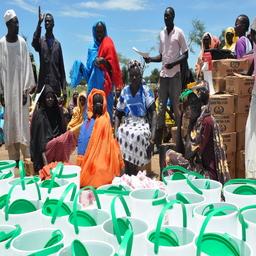}
\includegraphics [width=0.19\linewidth] {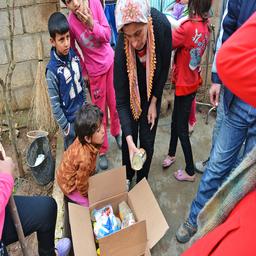}
\includegraphics [width=0.19\linewidth] {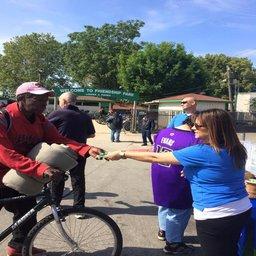}
\end{figure}

\noindent\rule{\textwidth}{1pt}
\newpage
\noindent\textbf{Verb:} EXAMINING \\
\noindent\textbf{Roles:} agent, place, item, tool
\begin{figure}[h]
\centering
\includegraphics [width=0.19\linewidth] {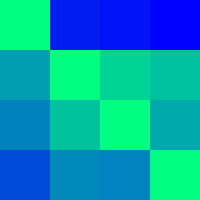}
\includegraphics [width=0.19\linewidth] {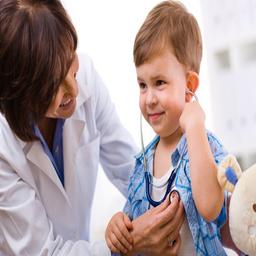}
\includegraphics [width=0.19\linewidth] {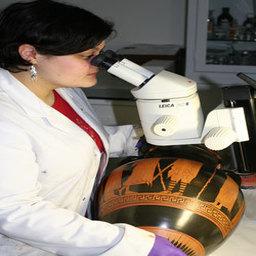}
\includegraphics [width=0.19\linewidth] {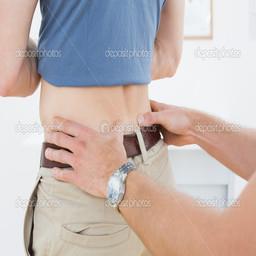}
\includegraphics [width=0.19\linewidth] {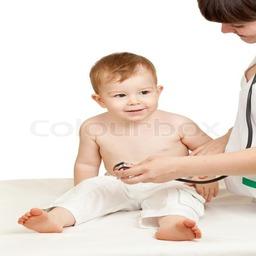}
\end{figure}

\noindent\rule{\textwidth}{1pt}

\noindent\textbf{Verb:} GIVING \\
\noindent\textbf{Roles:} agent, place, item, recipient
\begin{figure}[h]
\centering
\includegraphics [width=0.19\linewidth] {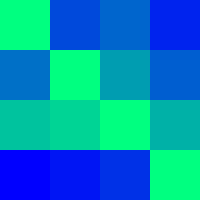}
\includegraphics [width=0.19\linewidth] {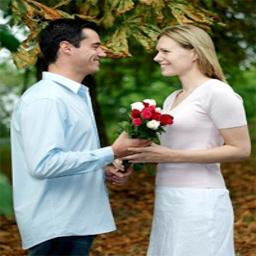}
\includegraphics [width=0.19\linewidth] {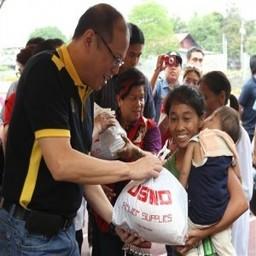}
\includegraphics [width=0.19\linewidth] {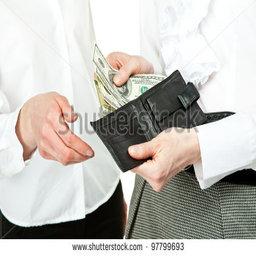}
\includegraphics [width=0.19\linewidth] {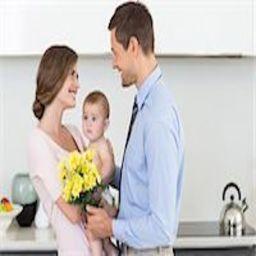}
\end{figure}

\noindent\rule{\textwidth}{1pt}
\noindent\textbf{Verb:} HUNCHING \\
\noindent\textbf{Roles:} agent, place, surface
\begin{figure}[h]
\centering
\includegraphics [width=0.19\linewidth] {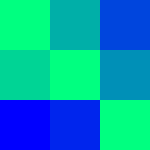}
\includegraphics [width=0.19\linewidth] {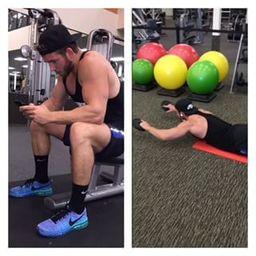}
\includegraphics [width=0.19\linewidth] {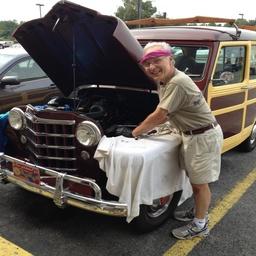}
\includegraphics [width=0.19\linewidth] {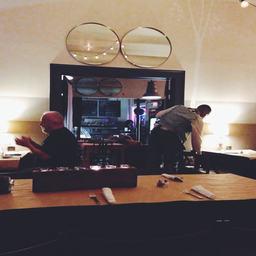}
\includegraphics [width=0.19\linewidth] {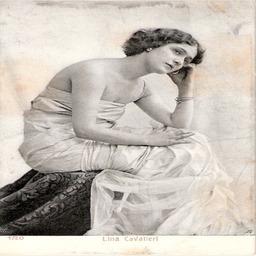}
\end{figure}

\noindent\rule{\textwidth}{1pt}

\noindent\textbf{Verb:} KISSING \\
\noindent\textbf{Roles:} agent, place, coagent, coagentpart, agentpart
\begin{figure}[h]
\centering
\includegraphics [width=0.19\linewidth] {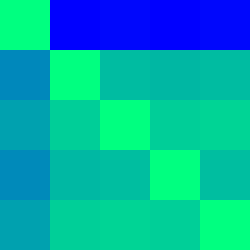}
\includegraphics [width=0.19\linewidth] {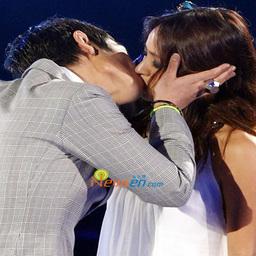}
\includegraphics [width=0.19\linewidth] {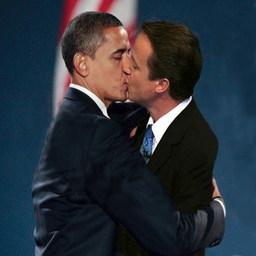}
\includegraphics [width=0.19\linewidth] {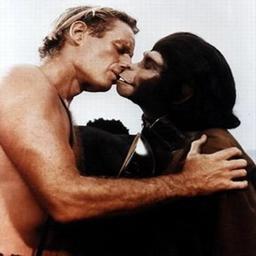}
\includegraphics [width=0.19\linewidth] {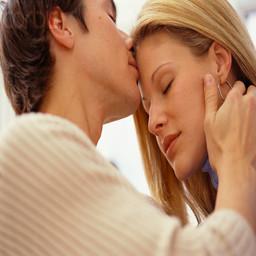}
\end{figure}

\noindent\rule{\textwidth}{1pt}
\newpage
\noindent\textbf{Verb:} MILKING \\
\noindent\textbf{Roles:} agent, place, source, tool, destination
\begin{figure}[h]
\centering
\includegraphics [width=0.19\linewidth] {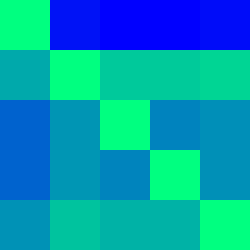}
\includegraphics [width=0.19\linewidth] {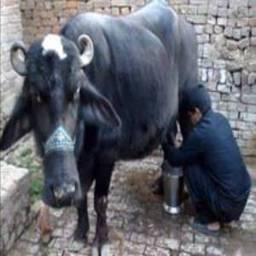}
\includegraphics [width=0.19\linewidth] {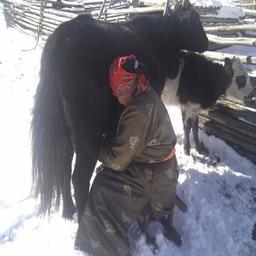}
\includegraphics [width=0.19\linewidth] {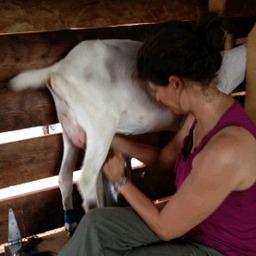}
\includegraphics [width=0.19\linewidth] {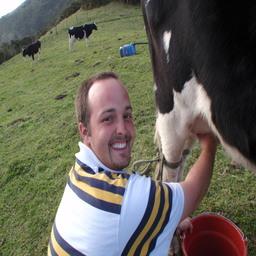}
\end{figure}

\noindent\rule{\textwidth}{1pt}

\noindent\textbf{Verb:} PACKING \\
\noindent\textbf{Roles:} agent, place, item, container
\begin{figure}[h]
\centering
\includegraphics [width=0.19\linewidth] {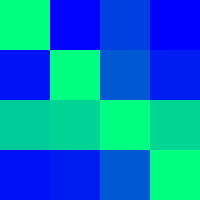}
\includegraphics [width=0.19\linewidth] {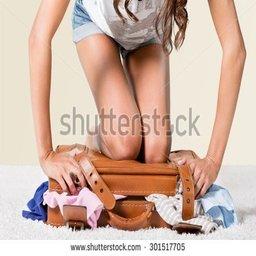}
\includegraphics [width=0.19\linewidth] {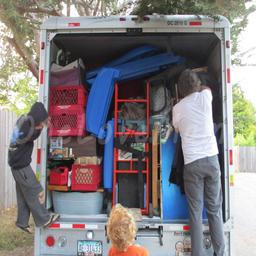}
\includegraphics [width=0.19\linewidth] {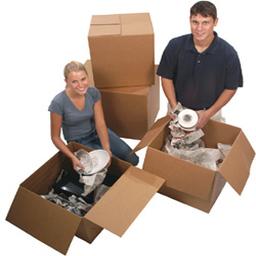}
\includegraphics [width=0.19\linewidth] {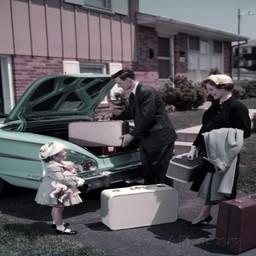}
\end{figure}

\noindent\rule{\textwidth}{1pt}
\noindent\textbf{Verb:} PERFORMING \\
\noindent\textbf{Roles:} agent, place, event, stage, tool
\begin{figure}[h]
\centering
\includegraphics [width=0.19\linewidth] {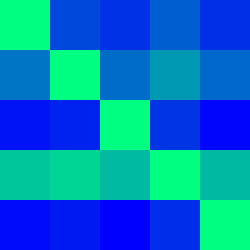}
\includegraphics [width=0.19\linewidth] {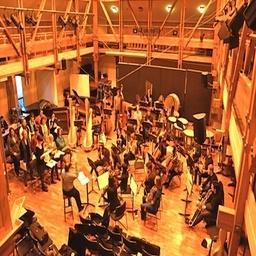}
\includegraphics [width=0.19\linewidth] {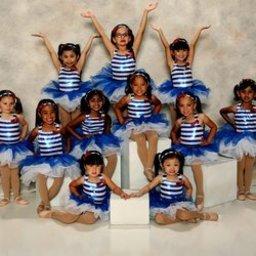}
\includegraphics [width=0.19\linewidth] {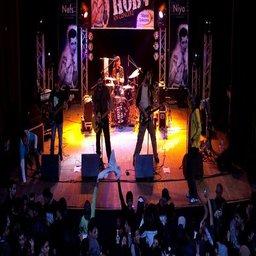}
\includegraphics [width=0.19\linewidth] {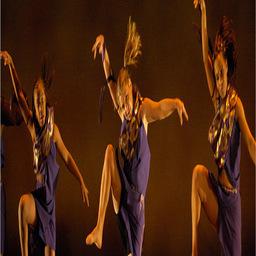}
\end{figure}

\noindent\rule{\textwidth}{1pt}